\documentclass[11pt,letterpaper]{article}
\usepackage[T1]{fontenc}
\usepackage{lmodern}
\usepackage[letterpaper,margin=0.87in,top=0.79in,bottom=0.83in]{geometry}
\ifdefined\pdftexversion
  \usepackage{microtype}
\fi
\usepackage{booktabs,multirow,graphicx,makecell}
\usepackage{amsmath,amsfonts,bm}

\def\eqref#1{equation~\ref{#1}}
\def\1{\bm{1}}

\DeclareMathAlphabet{\mathsfit}{\encodingdefault}{\sfdefault}{m}{sl}
\SetMathAlphabet{\mathsfit}{bold}{\encodingdefault}{\sfdefault}{bx}{n}

\usepackage[table]{xcolor}
\usepackage{pict2e}
\usepackage{amssymb,amsthm,algorithmic,algorithm}
\usepackage{adjustbox,tabularx,longtable}
\usepackage[font=small,labelfont=bf]{caption}
\usepackage{placeins}
\usepackage[authoryear,round]{natbib}
\setcitestyle{authoryear,round,citesep={;},aysep={,},yysep={;}}
\usepackage[hidelinks]{hyperref}

\ifdefined\pdfminorversion
\fi

\newtheorem{theorem}{Theorem}
\AtBeginDocument{\setlength{\bibsep}{3pt plus 1pt}}

\definecolor{frontmatterbg}{HTML}{F4F3F0}
\definecolor{frontmatterborder}{HTML}{DCD9D2}
\definecolor{resultshade}{HTML}{EDEAE3}
\newsavebox{\frontmatterbox}
\newlength{\frontmatterwidth}
\newlength{\frontmatterheight}
\makeatletter
\newcommand{\frontmatterbackground}{%
  \begingroup
  \setlength{\unitlength}{1pt}%
  \edef\panelright{\strip@pt\dimexpr\frontmatterwidth-16pt\relax}%
  \edef\paneltop{\strip@pt\dimexpr\frontmatterheight-16pt\relax}%
  \color{frontmatterbg}%
  \begin{picture}(0,0)
    \put(16,0){\rule{\dimexpr\frontmatterwidth-32pt\relax}{\frontmatterheight}}
    \put(0,16){\rule{\frontmatterwidth}{\dimexpr\frontmatterheight-32pt\relax}}
    \put(16,16){\circle*{32}}
    \put(\panelright,16){\circle*{32}}
    \put(16,\paneltop){\circle*{32}}
    \put(\panelright,\paneltop){\circle*{32}}
  \end{picture}%
  \endgroup
}
\makeatother
\newenvironment{frontmatter}{%
  \par\noindent
  \setlength{\frontmatterwidth}{\dimexpr\linewidth+32pt\relax}%
  \begin{lrbox}{\frontmatterbox}%
  \begin{minipage}{\dimexpr\frontmatterwidth-32pt\relax}%
  \setlength{\parskip}{5pt}%
}{%
  \end{minipage}%
  \end{lrbox}%
  \setlength{\frontmatterheight}{\dimexpr\ht\frontmatterbox+\dp\frontmatterbox+32pt\relax}%
  \makebox[\linewidth][c]{%
    \rlap{\raisebox{\dimexpr-16pt-\dp\frontmatterbox\relax}[0pt][0pt]{\frontmatterbackground}}%
    \kern16pt
    \raisebox{0pt}[\dimexpr\ht\frontmatterbox+16pt\relax][\dimexpr\dp\frontmatterbox+16pt\relax]{%
      \usebox{\frontmatterbox}}%
    \kern16pt
  }%
  \par\medskip
}

\makeatletter
\renewcommand{\maketitle}{\@maketitle}
\renewcommand{\@maketitle}{%
  \par\noindent
  {\raggedright\sffamily\bfseries\fontsize{17}{20}\selectfont
   \hbox{\raisebox{-6pt}[0pt][0pt]{\includegraphics[width=26pt]{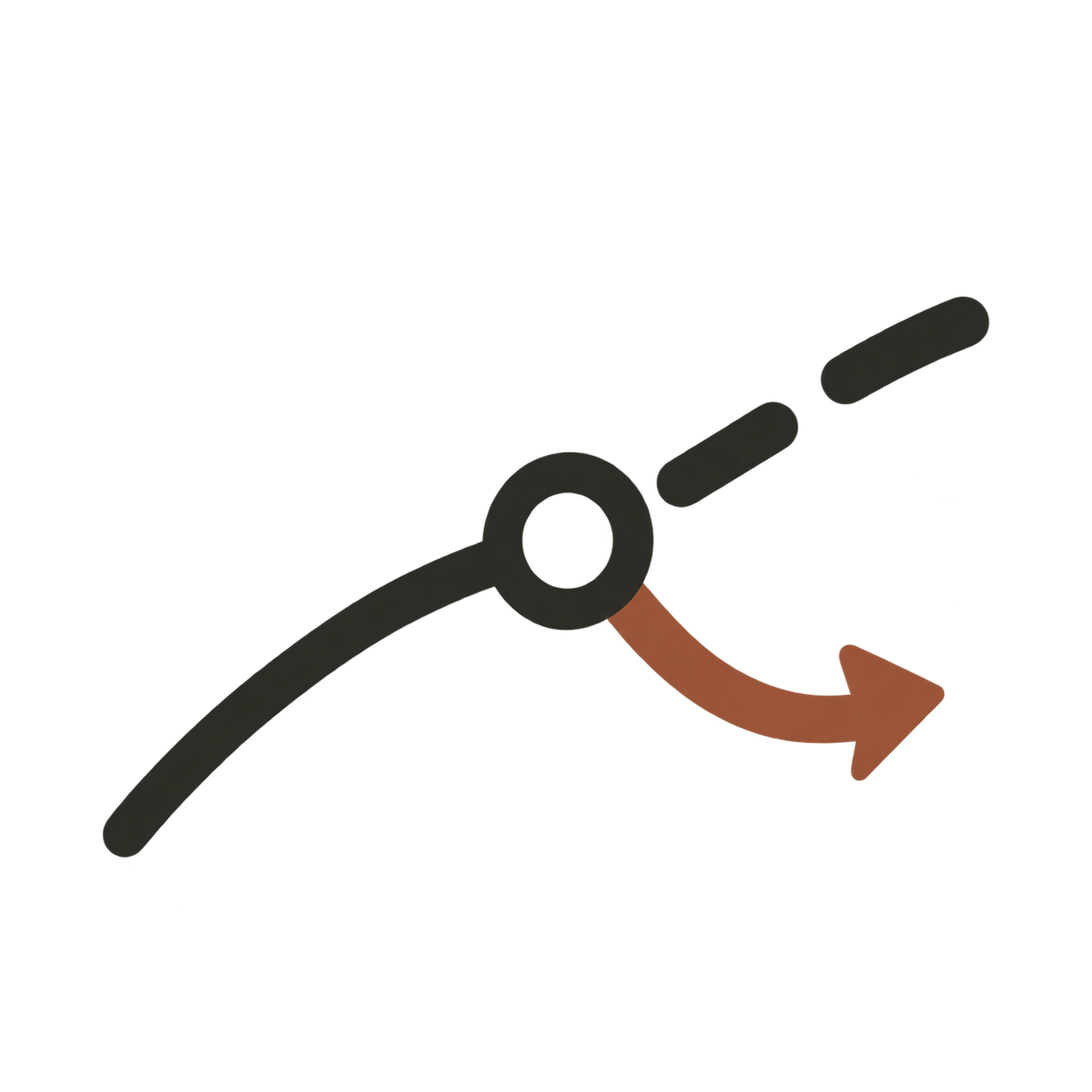}}\kern 0.2em}%
   \nobreak\@title\par}%
  \vspace{0.65em}%
  {\raggedright\@author\par}%
}
\renewcommand{\section}{\@startsection{section}{1}{\z@}%
  {-2.4ex plus -.7ex minus -.2ex}{1.1ex plus .2ex}%
  {\normalfont\Large\sffamily\bfseries\raggedright}}
\renewcommand{\subsection}{\@startsection{subsection}{2}{\z@}%
  {-2ex plus -.5ex minus -.2ex}{.7ex plus .2ex}%
  {\normalfont\large\bfseries\raggedright}}
\makeatother
\renewenvironment{abstract}{%
  \par\addvspace{0.7em}%
  \noindent{\color{frontmatterborder}\rule{\linewidth}{0.6pt}}\par
  \noindent{\sffamily\bfseries\abstractname}\par\nobreak\small\ignorespaces
}{\par}

\title{SAVE-WAM: Speculative Action Verification with Environment Feedback for Asynchronous World-Action Models}
\author{%
  {\sffamily\bfseries\small Ruiqi Xie$^{1}$, Tianchen Zhao$^{1}$, Yiran Shi$^{1}$, 
Feng Gao$^{1,2}$, Chao Yu$^{1}$, Qingmin Liao$^{1}$, Yu Wang$^{1}$\endgraf}
  \vspace{0.22em}
  {\footnotesize $^{1}$Tsinghua University \endgraf
    $^{2}$Striding AI\endgraf}
}
\date{}
\hypersetup{
  pdftitle={SAVE-WAM: Speculative Action Verification with Environment Feedback for Asynchronous World-Action Models},
  pdfauthor={Ruiqi Xie, Tianchen Zhao, Yiran Shi, Feng Gao, Chao Yu, Qingmin Liao, Yu Wang}
}

\begin{document}

\begin{frontmatter}
\maketitle

\begin{abstract}
Large generative robot policies often adopt asynchronous inference to reduce execution latency, but this introduces generation discrepancy: actions are generated from observations obtained before all previously predicted actions have been executed. Although existing methods attempt to mitigate this discrepancy through calibration during action generation, there lacks feedback from newer observations affected by ongoing execution. As a result, action generation remains \textbf{``open-looped''}, allowing accumulated prediction errors and eventually causing the trajectory to deviate from the synchronous inference. Therefore, we aim to explicitly model and correct the discrepancy introduced by asynchronous action prediction in a \textbf{``closed-looped''} manner.
Inspired by speculative decoding in language models, we introduce \textbf{SAVE-WAM}, a training-free post-generation feedback mechanism for asynchronous World-Action Models. SAVE-WAM treats asynchronously generated actions as speculative drafts and verifies them by measuring their deviation from the corresponding synchronous trajectory. Using updated observations during execution, the verifier detects excessively deviated trajectory, and triggers the replanning. As a plug-in mechanism, SAVE-WAM can be integrated with existing asynchronous inference methods to correct generation discrepancies with modest overhead, thereby improving the accuracy-efficiency trade-off. On LingBot-VA with FDM, SAVE-WAM improves success over unverified
asynchronous execution from $2\%$ to $42\%$ on \textit{Hanging Mug}
and from $82.76\%$ to $92.76\%$ overall, while retaining $73.8\%$
latency overlap. Moreover, verification enables the use of up to 1.97$\times$ larger action chunks without degrading task success, providing an additional dimension for improving generation efficiency, achieving up to $1.55\times$ end-to-end speedup on real-robot tasks while matching the success rate of synchronous execution.

\end{abstract}
\end{frontmatter}
\suppressfloats[t]
% ===================== Teaser Figure =====================
\begin{figure}[!htbp]
    \centering
    \includegraphics[width=0.94\linewidth]{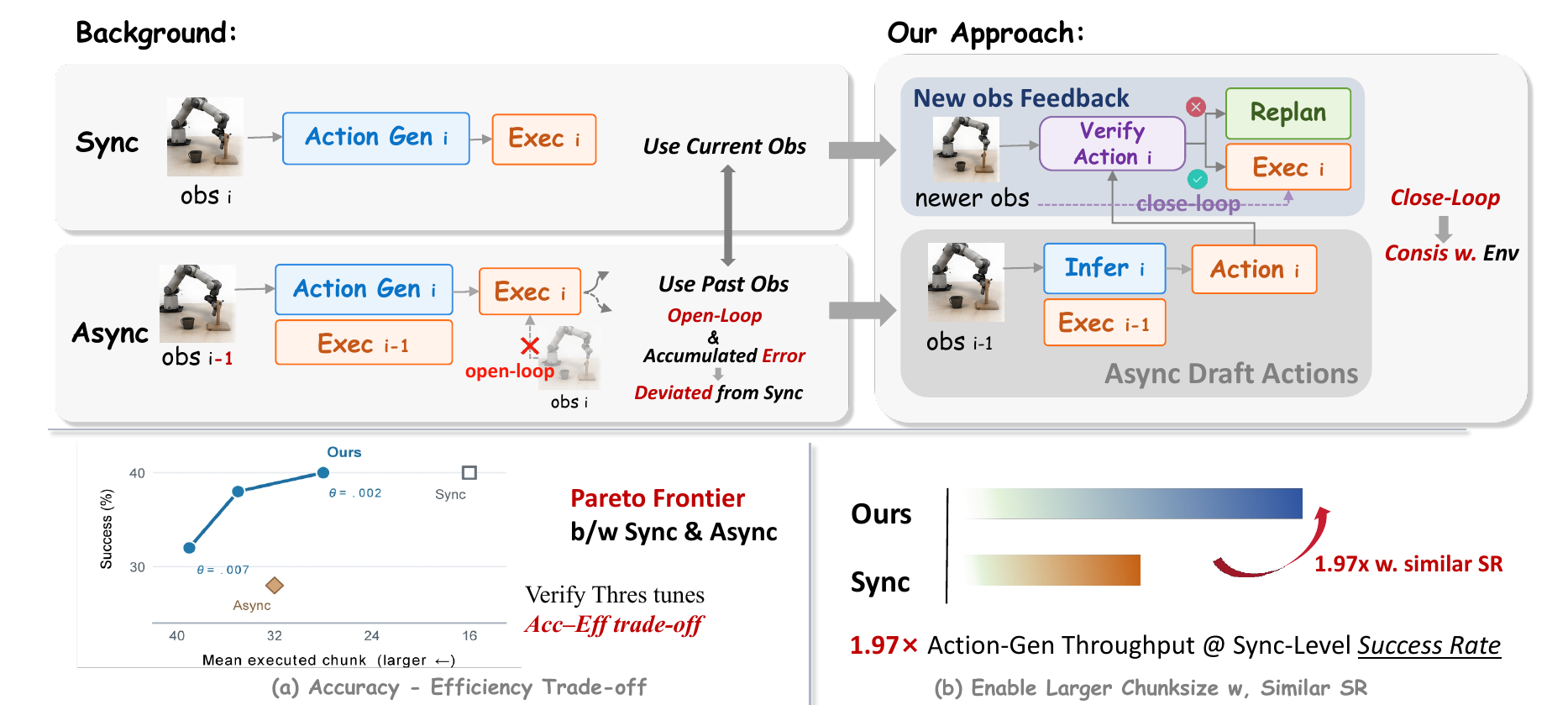}
    \caption{
    \textbf{Overview of SAVE-WAM.}
    Asynchronous inference overlaps action generation with execution to reduce latency, but actions are generated from past observations and remain open-loop with newer observation, causing trajectory deviation and accumulated error.
    SAVE-WAM treats asynchronously generated actions as speculative drafts and restores closed-loop feedback by verifying them with updated observations.
    This training-free plug-in improves the accuracy--efficiency trade-off, while enabling larger action chunks and up to $1.97\times$ higher action-generation throughput.
    }
    \label{fig:teaser}
\end{figure}
\section{Introduction}
\label{sec:introduction}

% Recent World Action Models (WAMs)~\citep{li2026causal, yuan2026fast} show robust ability in robot manipulation. However, generating an action requires multiple denoising steps to the model, which prolongs the task execution time, hindering realtime execution and practical deployment. In the standard synchronous inference scheme, the actions are generated based on the robot observation, then the next iteration's oberservation is taken after all predicated action execution. To reduce excution latency, the asynchronous inference scheme~\citep{black2026real, jiang2026futurertc} overlaps the action generation and excution time through advancing the next iteration's prediction before all previous action's execution. Despite providing speedup, the async inference involves action trajectory discrepancy compared with the accurate sync inference counterpart, for that it uses older observation before all previous actions are executed, and didnot directly take the effect of remaining action on the observaiton into account. 

Recent World Action Models (WAMs)~\citep{li2026causal, yuan2026fast,pai2025mimic,ye2026world,bi2026motus} have demonstrated strong performance in robotic manipulation. However, generating an action typically requires multiple denoising steps~\citep{akbari2026flash,zhao2026faster}, leading to substantial inference latency that prolongs task execution and hinders real-time control and practical deployment.
Under the standard synchronous inference scheme, actions are generated from the current robot observation, and the observation for the next iteration is acquired only after all predicted actions have been executed. To reduce execution latency, asynchronous inference methods~\citep{black2026real, jiang2026futurertc} overlap action generation with action execution by starting the next round of prediction before the previously generated actions have finished executing.
Although this overlap improves efficiency, asynchronous inference introduces a discrepancy in the predicted action trajectory relative to its synchronous counterpart~\citep{team2026world}. Specifically, the next action chunk is generated from an outdated observation captured before all previously predicted actions have been executed, and therefore does not directly account for the effects of the remaining actions on the robot state and subsequent observations.

Existing asynchronous methods mitigate the generation\textendash{}execution mismatch by improving future action prediction during generation, for example by conditioning on previously generated actions or explicitly predicting future observations~\citep{black2026real,jiang2026futurertc,li2026causal}. However, a direct source of feedback, the updated observations available after action generation and during execution\textemdash{}remains under-exploited. As a result, asynchronous inference still operates in an  ``open-loop'' manner: even small discrepancies in the predicted action trajectory can accumulate over time, eventually leading to substantial deviation from the synchronous reference trajectory.

To address this limitation, we aim to incorporate real-world feedback into asynchronous inference, thereby turning it into a ``closed-loop'' process. Inspired by Speculative Decoding~\cite{xia2023speculative}, we propose \textbf{SAVE-WAM}, which leverages updated observations acquired during action generation as feedback to dynamically adjust the inference process. Specifically, asynchronously generated actions are treated as drafts and verified by measuring their deviation from the corresponding synchronous reference trajectory. Rejected action tokens are discarded, triggering a replanning stage that regenerates a new action chunk conditioned on the latest observation.
We carefully design each component to minimize the overhead introduced by this additional verification. In particular, the feedback (verification) module is lightweight: it performs one-step flow scoring to compare the current draft trajectory against the reference trajectory using only a single diffusion step. Furthermore, we introduce an ahead-verification scheme that overlaps verification with action generation, further hiding the verification cost and minimizing the overall inference overhead.

The SAVE-WAM draft-and-verification scheme can serve as a plug-in module for existing asynchronous inference methods, compensating for their success-rate degradation with controllable and modest overhead. In this way, it turns a single operating point into a Pareto frontier over the accuracy\textendash{}efficiency trade-off. Furthermore, by using verification to prevent trajectory deviation, it can maintain task performance at larger action chunk sizes. This enables greater parallelism in action generation and, consequently, improves action-generation throughput.
Our contributions as follows:
\begin{itemize}
\item We introduce real-world feedback into asynchronous inference through a draft-and-verify scheme, and design a lightweight feedback module together with an ahead-verification mechanism to minimize the overhead.

\item SAVE-WAM is compatible with existing asynchronous inference methods, providing adjustable trade-off. Moreover, it supports larger action chunk sizes, enabling greater parallelism and higher action-generation throughput.

\item While achieving success rate comparable to that of synchronous baselines, SAVE-WAM preserves most of the execution-generation overlap enabled by Lingbot FDM, also increasing the feasible action chunk size by up to 1.97$\times$. In real-robot tasks, SAVE-WAM achieves up to 1.55$\times$ end-to-end speed up while matching the success rate of synchronous execution.
\end{itemize}

\section{Related Work}
\label{sec:related}
\paragraph{World-Action Models}
World-Action Models (WAMs) extend robot policies by predicting actions together with future observations, conditioning on current observations. Mimic-Video~\citep{pai2025mimic}, LingBot-VA~\citep{li2026causal}, DreamZero~\citep{ye2026world}, Cosmos Policy~\citep{kim2026cosmos}, and DiT4DiT~\citep{ma2026dit4dit} use imagined visual futures to inform action generation. Motus~\citep{bi2026motus} jointly generates video and actions, whereas FastWAM~\citep{yuan2026fast} and GigaWorld-Policy~\citep{ye2026gigaworld} implicitly use future foresight to generate actions during the reasoning stage for efficient action prediction. These works improve how actions are generated. We study what happens afterward: whether a generated action remains valid when a new real observation arrives.

\paragraph{Asynchronous Robot Policy Deployment}
Asynchronous inference generates actions while earlier actions execute, reducing stalls caused by large policy models. RTC~\citep{black2026real} maintains continuity through committed action prefixes. VLASH~\citep{tang2025vlash} and FutureRTC~\citep{jiang2026futurertc} train the model to enable it to generate actions in advance. WAM-based approaches often exploit explicit predictions of future observations to facilitate asynchronous inference, as in FDM~\citep{li2026causal} and AHA-WAM~\citep{cai2026aha}. These methods primarily improve the model\textquoteright{}s ability to generate actions asynchronously at generation time, whereas our approach is complementary: it can be combined with them to validate asynchronously pre-planned actions at execution time using feedback from updated observations.

\paragraph{Adaptive Action Execution}
Adaptive execution chooses when to stop following a planned chunk and replan. Prior methods mainly based on synchronous inference, using action uncertainty~\citep{liang2026adaptive,feng2026denoising}, internal model features~\citep{wang2026vla,xu2026knowing}, imagined--real observation consistency~\citep{wang2026trust,pan2026vla}, reinforcement learning signals~\citep{zhao2026dynamic}, or task progress~\citep{ye2026rethink}. We likewise use feedback to determine the executed horizon, but target asynchronous open-loop problem. A training-free verifier checks successive segments of a previously generated draft against updated observations, allowing valid actions to stream through and rejecting the remaining suffix when they become inconsistent.

\paragraph{Speculative Inference and Verification}
Speculative decoding~\citep{leviathan2023fast,xia2023speculative} accelerates generation by proposing drafts, accepting verified prefixes, and regenerating rejected suffixes. Related acceleration ideas extend to continuous generative models~\citep{de2025accelerated,pan2026specdiff}. We apply the draft-and-verify principle to robot execution, where validity depends on the changing physical environment feedback.

\section{Preliminaries}
\label{sec:preliminaries}

\paragraph{Feedback in robot execution.}
In feedback control~\citep{doyle2013feedback}, control inputs change the state of a plant, and the resulting measurements are used to update subsequent inputs. For a robot policy, the robot and its environment form the plant:
\begin{equation}
    x_{t+1}=f(x_t,a_t),
    \qquad
    o_{t+1}=q(x_{t+1}),
    \label{eq:robot-plant}
\end{equation}
where $a_t$ is the executed action, $x_t$ is the environment state, and $o_t$ is the observation provided to the policy as feedback. The policy maps an observation to a chunk of $H$ future actions,
\begin{equation}
    \mathbf{A}_t
    =
    \pi(o_t)
    =
    [a_t,\ldots,a_{t+H-1}].
    \label{eq:action-chunk-policy}
\end{equation}

In synchronous inference, the policy waits for the previous chunk to finish, observes the resulting state, and then generates the next chunk. This allows action generation to use the latest observation, but introduces an inference pause. Asynchronous inference instead generates actions before all previously predicted actions have been executed. It therefore lacks feedback from newer observations affected by ongoing execution, leaving the pre-generated actions effectively open-looped and allowing prediction errors to accumulate.

We describe such feedback through its \emph{input}, \emph{module}, and \emph{output}:
\begin{equation}
    z
    =
    \mathcal{F}(o^{\mathrm{new}}),
    \qquad
    \mathbf{A}^{\mathrm{exec}}
    =
    \mathcal{U}(\mathbf{A}^{\mathrm{pending}},z).
    \label{eq:feedback-interface}
\end{equation}
Here, $o^{\mathrm{new}}$ is the feedback input, $\mathcal{F}$ is the feedback module that produces a signal $z$, and $\mathcal{U}$ determines the feedback output by controlling which pending actions enter the execution queue.

\paragraph{Latency in chunked execution.}
Let $\Delta t$ be the controller period and $T_H=H\Delta t$ the execution duration of an $H$-action chunk. Let $T_{\mathrm{plan}}$ denote the latency of policy inference, and $T_{\mathrm{halt}}$ the uncovered waiting time between consecutive chunks. A synchronous policy places the inference latency on the execution critical path,
\begin{equation}
    T_{\mathrm{halt}}^{\mathrm{sync}}=T_{\mathrm{plan}},
\end{equation}
whereas asynchronous execution overlaps inference with ongoing execution,
\begin{equation}
    T_{\mathrm{halt}}^{\mathrm{async}}
    =
    \max(0,T_{\mathrm{plan}}-T_H).
    \label{eq:async-halt}
\end{equation}

A larger chunk improves efficiency in two ways. First, it provides a longer execution window to hide policy inference. Second, each policy call generates more executable actions, reducing the inference cost per action to approximately $T_{\mathrm{plan}}/H$. When $T_{\mathrm{plan}}$ grows slowly with $H$, increasing the chunk size can therefore reduce the effective inference cost per action and increase action-generation throughput. However, larger chunks also rely on older observations for longer, increasing the discrepancy between generation and execution. This efficiency--accuracy trade-off motivates adding feedback that enables larger chunks while correcting stale actions before execution.

\section{Method}
\label{sec:method}

\paragraph{Overview.}

We introduce a feedback-gated framework that introduces closed-loop feedback into asynchronous action execution. Compared with synchronous chunking, asynchronous inference generates actions further ahead of their execution, thereby extending the effective open-loop interval. Our method closes this loop through three components. Section~\ref{sec:feedback-output} defines the \emph{feedback input \& output}: how newer observations provide feedback to ongoing action execution. Section~\ref{sec:validity} defines the \emph{feedback module}: a low-cost, one-step flow score that determines whether the generated actions remain valid. Section~\ref{sec:ahead-verification} further reduces the overhead introduced by the feedback mechanism by delivering recent real observations to the feedback module without stalling action execution. The system can therefore accept more actions in predictable states while triggering earlier replanning when new feedback reveals a mismatch.

\begin{figure}[t]
    \centering
    \includegraphics[width=0.8\linewidth]{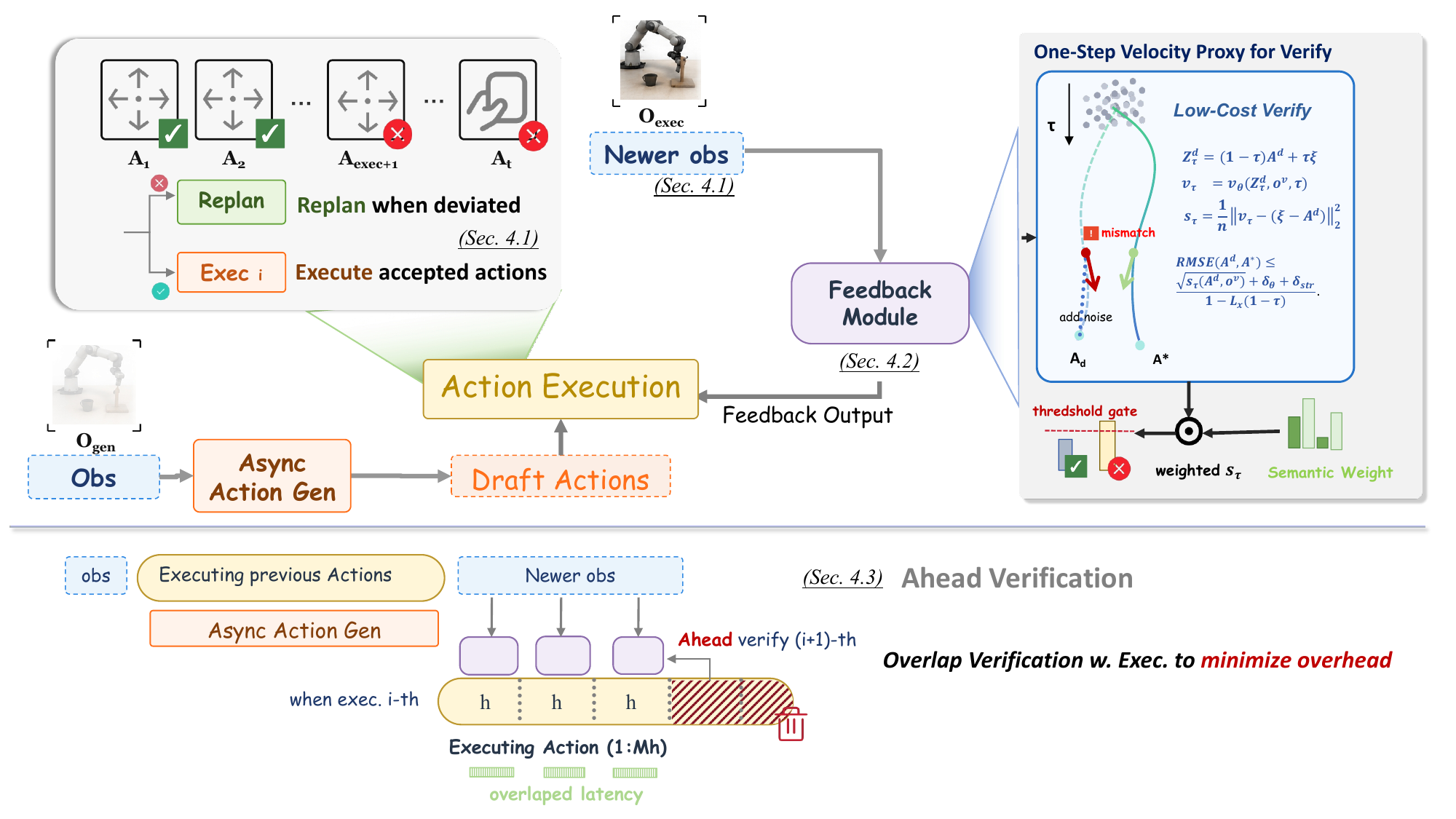}
    \caption{
    \textbf{Feedback mechanism of SAVE-WAM.}
    The policy generates speculative action chunks from a planning observation,
    while the feedback module decides whether pending actions remain executable
    under newer execution-time observations.
    Valid actions are committed to the queue; mismatched actions are discarded
    and followed by replanning, yielding a dynamic accepted prefix.
    }
    \label{fig:feedback}
\end{figure}

\subsection{Feedback-Gated Action Execution}
\label{sec:feedback-output}

Asynchronous inference lacks feedback from newer observations during execution, leaving pre-generated actions effectively open-looped. We therefore introduce \emph{feedback-gated action execution} that uses newer observations to verify draft actions, defining the feedback \emph{input} as the newer observation and the feedback \emph{output} as whether the pending actions should be executed.

Given a generation-time observation $o^{\mathrm{gen}}$, the asynchronous policy first produces a speculative draft:
\begin{equation}
    \mathbf{A}^{\mathrm d}
    =
    \pi(o^{\mathrm{gen}}),
\end{equation}
where $\pi$ denotes the underlying generative robot policy and
$\mathbf{A}^{\mathrm d}$ denotes its full draft action sequence.

As shown in Figure~\ref{fig:feedback}, we decouple action proposal from action commitment. The policy determines \emph{what} actions to propose, while feedback determines \emph{whether} draft actions remain valid under newer observations. Given a pending draft segment $A^{\mathrm d}$ and a newer observation $o^{\mathrm{v}}$, the verifier produces
\begin{equation}
    {
    g
    =
    \mathcal{G}(A^{\mathrm d},o^{\mathrm{v}})
    \in\{0,1\},
    \qquad
    A^{\mathrm{exec}}
    =
    \begin{cases}
        A^{\mathrm d}, & g=1,\\
        \varnothing, & g=0,
    \end{cases}
    }
    \label{eq:execution-time-gating}
\end{equation}
{
where $\mathcal{G}$ denotes the verification function and $g$ is its binary acceptance decision.
}
Accepted actions are executed, while rejected actions are discarded and replanned from a recent observation. Repeating this process turns a fixed draft horizon into a dynamic accepted prefix, allowing longer reuse in predictable states and earlier replanning when a mismatch occurs.

The remaining challenge is to provide this feedback with minimal overhead. Sections~\ref{sec:validity} and~\ref{sec:ahead-verification} reduce the verification cost and avoid stalling action execution, respectively.

\subsection{Low-Cost Draft Verification}
\label{sec:validity}

The feedback mechanism above requires determining whether a draft action segment should still be executed under a newer observation. Directly regenerating the corresponding actions provides an oracle verification signal, but largely removes the efficiency benefit of asynchronous execution. We therefore introduce a low-cost draft verification method.

\paragraph{Oracle verification.}
Let $A^{\mathrm d}\in\mathbb{R}^{h\times d_a}$ be an $h$-step draft segment awaiting verification, where $h$ is the segment length and $d_a$ is the action dimension, and let $o^{\mathrm v}$ be a newer observation used for verification. 

An oracle reruns the policy on $o^{\mathrm v}$, as in synchronous planning, and directly compares the draft segment with newly generated reference actions:
\begin{equation}
    A^\star=\pi_h(o^{\mathrm v};\xi),
    \qquad
    d^{\mathrm{oracle}}
    =
    \frac{\|A^{\mathrm d}-A^\star\|_2}{\sqrt{n}},
    \label{eq:oracle-validity}
\end{equation}
where $n=hd_a$, $\pi_h$ extracts the \emph{temporally aligned} $h$-step block from the full policy output, and
{
$\xi\sim\mathcal{N}(0,I_n)$ denotes the Gaussian flow noise used for the corresponding generation trajectory. 
}
A small discrepancy indicates that the draft remains close to the synchronous policy decision under the updated observation. However, computing $A^\star$ requires full multi-step inference at a cost comparable to replanning.

\paragraph{One-step verification.}
We replace oracle regeneration with a single vector-field evaluation.
{
For verification, we use a fixed noise level $\tau$ and sample
$\xi\sim\mathcal{N}(0,I_n)$. We construct
}
\begin{equation}
    {
    Z_\tau^{\mathrm d}
    =
    (1-\tau)A^{\mathrm d}+\tau\xi,
    \qquad
    v_{\mathrm{target}}^{\mathrm d}
    =
    \xi-A^{\mathrm d}.
    }
    \label{eq:draft-flow-state}
\end{equation}
{
}

We then replace the original planning observation with $o^{\mathrm v}$ and evaluate the policy vector field once:
\begin{equation}
    {
    \hat v_\tau^{\mathrm d}
    =
    v_\theta(Z_\tau^{\mathrm d},o^{\mathrm v},\tau),
    \qquad
    r_\tau
    =
    \hat v_\tau^{\mathrm d}
    -
    v_{\mathrm{target}}^{\mathrm d},
    }
    \label{eq:one-step-residual}
\end{equation}
where $v_\theta$ is the learned action vector field and $r_\tau$ is the one-step velocity residual. The corresponding score that will compare with thredhold is
\begin{equation}
    s_\tau(A^{\mathrm d},o^{\mathrm v})
    =
    \frac{1}{n}\|r_\tau\|_2^2.
    \label{eq:one-step-validity-score}
\end{equation}
If the draft remains compatible with the updated observation, the predicted velocity should stay close to the target velocity associated with the noised draft. This provides a low-cost verification signal with only one vector-field evaluation, without generating a new action trajectory.

\begin{theorem}[Action-error bound for one-step verification]
\label{thm:one-step-validity}
Let $v^\star$ denote the target vector field conditioned on $o^{\mathrm v}$.
Assume that the model approximation error is bounded by $\delta_\theta$,
$v^\star$ is $L_x$-Lipschitz with respect to the action state, and its
deviation from the linear flow between $A^\star$ and $\xi$ is bounded by
$\delta_{\mathrm{str}}$. For any $\tau$ satisfying
$L_x(1-\tau)<1$,
\begin{equation}
    d^{\mathrm{oracle}}
    \leq
    \frac{
        \sqrt{s_\tau(A^{\mathrm d},o^{\mathrm v})}
        +\delta_\theta
        +\delta_{\mathrm{str}}
    }{
        1-L_x(1-\tau)
    }.
    \label{eq:one-step-action-bound}
\end{equation}
\end{theorem}

Theorem~\ref{thm:one-step-validity} shows that the canonical one-step residual controls the same action discrepancy between the draft and synchronous reference measured by the oracle. Its proof is given in Appendix~\ref{app:one-step-validity-proof}, and their empirical correlation is further evaluated in Appendix~\ref{sec:one-step-exp}.
{
Empirically, the one-step verifier achieves AUROC values of $0.896$ and $0.900$ against the full-denoising oracle, providing direct evidence for the approximation gate.
}

{
\paragraph{Practical verification rule.}
Position, rotation, and gripper dimensions of the action have different semantics and residual scales. We therefore partition the action dimensions into
$k\in\{p,r,g\}$, denoting position, rotation, and gripper, respectively.
Let $r_{\tau,k}$ be the residual restricted to group $k$, and let $n_k$ denote the number of scalar elements in that group over the $h$-step segment. We compute
\begin{equation}
    e_k
    =
    \frac{1}{n_k}
    \|r_{\tau,k}\|_2^2,
    \qquad k\in\{p,r,g\},
    \label{eq:group-verification-error}
\end{equation}
and use the weighted score
\begin{equation}
    S(A^{\mathrm d},o^{\mathrm v})
    =
    0.65e_p+0.20e_r+0.15e_g.
    \label{eq:weighted-verification-score}
\end{equation}
The weighting balances the contributions of action groups with heterogeneous numerical scales, preventing one group from dominating the decision and empirically allowing longer accepted action horizons. Full implementation details and the corresponding analysis are provided in Appendix~\ref{app:verifier-details}.
}

{
We use a fixed verification threshold of $0.004$ for all tasks within each backbone/benchmark configuration, and further discuss the choices of the threshold and noise level $\tau=0.3$ in Appendix~\ref{app:verifier-details}.
}

\subsection{Speculative Ahead Verification}
\label{sec:ahead-verification}

Although the verifier in Section~\ref{sec:validity} greatly reduces verification cost, performing verification immediately before execution places its latency $T_{\mathrm{ver}}$ on the execution critical path. However, since verification is now much cheaper than full action generation, it only needs to be moved a few actions ahead to hide this latency. We therefore verify upcoming actions while the current ones are still being executed, using a recent observation and finishing before the verified actions are needed for execution.

\paragraph{Large proposal blocks, small verification segments.}
We preserve the long generation horizon of asynchronous inference, but partition each draft into $M$ consecutive segments:
\begin{equation}
    \mathbf{A}^{\mathrm d}
    =
    [A_1^{\mathrm d},\ldots,A_M^{\mathrm d}],
    \qquad
    H=Mh,
    \label{eq:speculative-blocks}
\end{equation}
where $H$ is the full proposal horizon and each segment
$A_j^{\mathrm d}$ contains $h$ actions. The large generation horizon $H$ reduces how often full policy inference is needed, while the smaller segment size $h$ enables more frequent feedback within the generated draft.

{
When $A_j^{\mathrm d}$ has been committed for execution, the robot acquires a newer observation $o_j^{\mathrm v}$ and verifies the next segment $A_{j+1}^{\mathrm d}$ in parallel. To preserve temporal alignment, we form a $2h$-step verification window
\begin{equation}
    \widetilde A_j^{\mathrm d}
    =
    [A_j^{\mathrm d},A_{j+1}^{\mathrm d}]
    \in\mathbb{R}^{2h\times d_a},
\end{equation}
and evaluate the vector field over the entire window under $o_j^{\mathrm v}$. Since the first $h$ actions correspond to the already committed segment, only the velocity residual over the aligned second block, with relative indices $[h,2h)$, is used to compute the verification score for $A_{j+1}^{\mathrm d}$:
\begin{equation}
    g_{j+1}
    =
    \mathcal{G}\left(
        A_{j+1}^{\mathrm d},
        o_j^{\mathrm v}
    \right).
    \label{eq:one-block-ahead}
\end{equation}
Thus, each ahead-verification call evaluates velocities over $2h$ actions for temporal alignment, while the feedback decision is determined only by the target $h$-step segment. Further implementation details and pseudocode are provided in Appendix~\ref{app:ahead-alignment}.
}

Compared with the observation used to generate $\mathbf{A}^{\mathrm d}$, $o_j^{\mathrm v}$ is much closer to the execution time of $A_{j+1}^{\mathrm d}$. Since the current segment provides an execution window
\begin{equation}
    T_h=h\Delta t,
\end{equation}
where $\Delta t$ is the controller period, verification introduces no additional stall when
\begin{equation}
    T_{\mathrm{ver}}\leq T_h.
    \label{eq:ahead-deadline}
\end{equation}

\paragraph{Pipelined feedback and replanning.}
If $g_{j+1}=1$, the next segment is accepted and executed immediately after the current one, while verification moves on to the following segment. If $g_{j+1}=0$, the remaining unexecuted actions
$[A_{j+1}^{\mathrm d},\ldots,A_M^{\mathrm d}]$ are discarded and the policy replans from a recent observation. In this way, low-cost verification runs in parallel with action execution, minimizing the additional overhead introduced by feedback. This largely maintains the latency benefit of asynchronous inference while allowing newer observations to provide feedback.

\section{Experiments}
\subsection{Experimental Setup}
\label{sec:experimental-setup}

\begin{table}[!htbp]
\centering
\caption{Results on RoboTwin for two policy backbones. E and H denote the clean (easy) and random (hard) settings. SR is success rate (\%), Chunk is the executed chunk size, and Time is the average completion time (s) over successful episodes. $T_{\mathrm{inf}}$ is the average total model-inference time (s) over successful episodes. Full results are in Appendix~\ref{app:full_task_results}.}
\label{tab:robotwin-two-models}
\begingroup
\scriptsize
\setlength{\tabcolsep}{3pt}
\renewcommand{\arraystretch}{1.04}
\setlength{\aboverulesep}{0pt}
\setlength{\belowrulesep}{0pt}
\newcommand{\ourscell}[1]{\cellcolor{resultshade}#1}
\newcommand{\tightellipsis}{%
  \noalign{\vskip 0.5pt
    \hbox to \linewidth{\hfil\raisebox{-2pt}[0pt][0pt]{{\boldmath$\cdot\mkern1mu\cdot\mkern1mu\cdot$}}\hfil}%
    \vskip 3.5pt}}
\newcolumntype{N}{>{\centering\arraybackslash}p{0.075\linewidth}}
\newcolumntype{Q}{>{\centering\arraybackslash}p{0.105\linewidth}}

% =========================================================
% LingBot-VA
% T_inf is taken from the clean sheet.
% =========================================================
\begin{tabularx}{\linewidth}{@{}p{0.17\linewidth}>{\raggedright\arraybackslash}X N N N N Q Q@{}}
\toprule
\rowcolor{black!5}
\multicolumn{8}{c}{\rule{0pt}{2.2ex}\small\bfseries LingBot-VA} \\
\midrule
\multirow{2}{*}{Task} & \multirow{2}{*}{Baseline}
& \multicolumn{2}{c}{SR (\%) $\uparrow$}
& \multicolumn{2}{c}{Chunk $\uparrow$}
& \multirow{2}{*}{Time (s) $\downarrow$}
& \multirow{2}{*}{$T_{\mathrm{inf}}$ (s) $\downarrow$} \\
\cmidrule(lr){3-4}\cmidrule(lr){5-6}
& & E & H & E & H & & \\
\midrule

\multirow{5}{*}{Hanging Mug}
& \ourscell{Ours (Async-64+Verify)}
& \ourscell{\textbf{38}}
& \ourscell{42}
& \ourscell{35.03}
& \ourscell{28.03}
& \ourscell{139.00}
& \ourscell{47.09} \\
& FDM Async-32 & 26 & 30 & 32.00 & 32.00 & 128.69 & 50.52 \\
& FDM Async-64 & 4 & 2 & 64.00 & 64.00 & 121.28 & 25.77 \\
& Sync-16 & 30 & \textbf{46} & 16.00 & 16.00 & 156.61 & 89.61 \\
& AAC & 20 & 16 & 30.65 & 30.83 & 135.28 & 53.82 \\

\midrule

\multirow{5}{*}{Place Can Basket}
& \ourscell{Ours (Async-64+Verify)}
& \ourscell{82}
& \ourscell{\textbf{88}}
& \ourscell{34.70}
& \ourscell{29.02}
& \ourscell{123.53}
& \ourscell{37.49} \\
& FDM Async-32 & 74 & 80 & 32.00 & 32.00 & 113.82 & 39.84 \\
& FDM Async-64 & 76 & 80 & 64.00 & 64.00 & 105.45 & 20.33 \\
& Sync-16 & \textbf{84} & 86 & 16.00 & 16.00 & 133.70 & 70.67 \\
& AAC & 70 & 80 & 30.19 & 30.34 & 123.94 & 43.09 \\

\tightellipsis
\midrule

\multirow{5}{*}{Avg. (50 tasks)}
& \ourscell{Ours (Async-64+Verify)}
& \ourscell{92.76}
& \ourscell{89.20}
& \ourscell{31.55}
& \ourscell{28.41}
& \ourscell{91.43}
& \ourscell{30.99} \\
& FDM Async-32 & 85.64 & 83.68 & 32.00 & 32.00 & 80.31 & 30.99 \\
& FDM Async-64 & 82.76 & 83.20 & 64.00 & 64.00 & 76.66 & 15.81 \\
& Sync-16 & \textbf{92.80} & \textbf{90.68} & 16.00 & 16.00 & 97.97 & 54.97 \\
& AAC & 85.32 & 82.08 & 27.61 & 27.52 & 86.76 & 35.48 \\

\bottomrule
\end{tabularx}

\vspace{0.4em}

% =========================================================
% FastWAM
% T_inf is taken from the random sheet.
% =========================================================
\begin{tabularx}{\linewidth}{@{}p{0.17\linewidth}>{\raggedright\arraybackslash}X N N N N Q Q@{}}
\toprule
\rowcolor{black!5}
\multicolumn{8}{c}{\rule{0pt}{2.2ex}\small\bfseries FastWAM} \\
\midrule
\multirow{2}{*}{Task} & \multirow{2}{*}{Baseline}
& \multicolumn{2}{c}{SR (\%) $\uparrow$}
& \multicolumn{2}{c}{Chunk $\uparrow$}
& \multirow{2}{*}{Time (s) $\downarrow$}
& \multirow{2}{*}{$T_{\mathrm{inf}}$ (s) $\downarrow$} \\
\cmidrule(lr){3-4}\cmidrule(lr){5-6}
& & E & H & E & H & & \\
\midrule

\multirow{4}{*}{Move Stapler Pad}
& \ourscell{Ours (RTC + Verify)}
& \ourscell{\textbf{90}}
& \ourscell{\textbf{78}}
& \ourscell{18.73}
& \ourscell{18.40}
& \ourscell{34.02}
& \ourscell{8.69} \\
& RTC & 78 & 76 & 19.25 & 19.24 & 28.09 & 8.31 \\
& Sync-12 & 76 & 74 & 12.00 & 12.00 & 42.81 & 11.42 \\
& AAC & 86 & 68 & 16.12 & 15.94 & 34.36 & 10.03 \\

\midrule

\multirow{4}{*}{Put Object Cabinet}
& \ourscell{Ours (RTC + Verify)}
& \ourscell{\textbf{94}}
& \ourscell{\textbf{92}}
& \ourscell{19.48}
& \ourscell{18.61}
& \ourscell{56.20}
& \ourscell{14.29} \\
& RTC & 90 & 80 & 19.50 & 19.64 & 52.27 & 13.54 \\
& Sync-12 & 88 & 86 & 12.00 & 12.00 & 63.85 & 18.99 \\
& AAC & 90 & 90 & 16.55 & 15.95 & 60.86 & 16.68 \\

\tightellipsis
\midrule

\multirow{4}{*}{Avg. (50 tasks)}
& \ourscell{Ours (RTC + Verify)}
& \ourscell{\textbf{92.52}}
& \ourscell{\textbf{92.72}}
& \ourscell{18.97}
& \ourscell{18.28}
& \ourscell{41.02}
& \ourscell{10.94} \\
& RTC & 90.92 & 90.96 & 18.98 & 18.94 & 34.97 & 10.52 \\
& Sync-12 & 92.24 & 91.88 & 12.00 & 12.00 & 49.79 & 14.43 \\
& AAC & 91.72 & 91.76 & 16.50 & 15.89 & 43.72 & 12.67 \\

\bottomrule
\end{tabularx}
\endgroup
\end{table}

\paragraph{Backbones.}
We evaluate our method on two flow-based World Action Models with different
asynchronous mechanisms. LingBot-VA uses its native FDM to predict future
observations during asynchronous inference~\citep{li2026causal}, with 5
video and 10 action denoising steps. FastWAM is paired with RTC, which
conditions asynchronous generation on committed actions, with its default denoising steps~\citep{yuan2026fast,black2026real}.

\paragraph{Benchmarks.}
We evaluate all tasks in RoboTwin~2.0~\citep{chen2025robotwin} under both clean
and randomized settings, with 50 rollouts per task. On
LIBERO~\citep{liu2023libero}, we evaluate all tasks for FastWAM, using 50 rollouts per task. For LingBot-VA, we report results only on LIBERO-10, as the publicly released checkpoint is trained exclusively on this suite.

\paragraph{Baselines.}
We compare against synchronous execution, each backbone's native asynchronous scheme, and AAC adapted to asynchronous execution~\citep{liang2026adaptive}. Labels such as Async-64 denote asynchronous execution with an action chunk size of 64. Our method uses the same asynchronous backbones, adding pipelined verification and feedback-triggered replanning. {We further include EMA and BID\citep{liu2025bidirectional} as well as AutoHorizon~\citep{wang2026vla} and VLA-Corrector~\citep{pan2026vla} as closely related uncertainty- and correction-based methods. Their implementations and full LIBERO results are provided in Appendix~\ref{app:additional-baselines}.}

\paragraph{Metrics.}
We report \textbf{Success Rate}, \textbf{Average Executed Chunk Size}, \textbf{Inference Latency} ($T_{\mathrm{infer}}$), and \textbf{Task Completion Time}. Average Executed Chunk Size measures the number of actions executed per full policy inference and reflects action-generation throughput. $T_{\mathrm{infer}}$ measures the latency of policy inference. Task Completion Time measures the end-to-end wall-clock time, including inference and execution, and is averaged over successful rollouts.
In the main simulation experiments, simulator stepping dominates wall-clock time, so end-to-end timing only partially reflects the speedup from inference--execution overlap.

\subsection{Experiments Results}
\label{sec:experimental-results}

\paragraph{Recovering asynchronous success with verification overhead.}
Tables~\ref{tab:robotwin-two-models} and~\ref{tab:libero-model-separated} show that verification recovers much of the success lost in asynchronous execution, at the cost of extra verification and replanning. On RoboTwin, SAVE-WAM improves LingBot-VA over FDM Async-64 by 10.00 and 6.00 percentage points in the clean and random settings, while increasing completion time from 76.66 to 91.43s. {Nevertheless, $73.8\%$ of the accumulated policy inference time remains hidden by asynchronous execution, with detailed timing data in Appendix~\ref{app:timing-breakdown}.} For FastWAM, success improves by 1.60 and 1.76 points over RTC, with time increasing from 34.97 to 41.02s. On LIBERO-10, LingBot-VA improves from 69.40\% to 95.80\%, while time increases from 54.87 to 57.16s. Overall, post-generation feedback moves asynchronous execution toward higher-success operating points with moderate extra time cost.

\begin{figure}[!ht]
    \centering
    \includegraphics[width=0.46\linewidth]{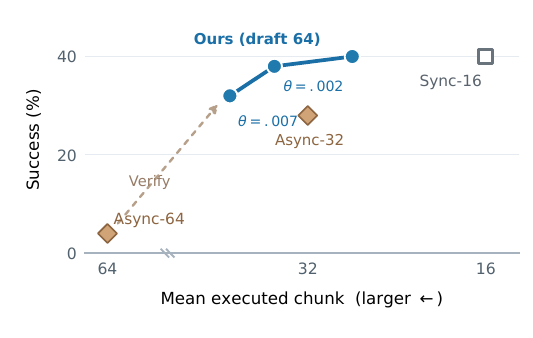}
    \caption{\textbf{Chunk-size--success trade-off on \textit{Hanging Mug}.}
    Varying the threshold $\theta$ recovers success from
    Async-64 and traces Pareto frontier.}
    \label{fig:hanging-mug-pareto}
\end{figure}

\paragraph{Tuning the success--efficiency trade-off.}
The verification threshold controls how much of a speculative draft is accepted before replanning, providing an adjustable trade-off between success and action reuse. As shown in Figure~\ref{fig:hanging-mug-pareto}, a threshold of 0.002 reaches 40\% success with an average executed chunk of 28. Increasing it to 0.004 gives 38\% success with a chunk size of 35, while 0.007 gives 32\% success with a chunk size of 39. Thus, stricter verification replans earlier and favors success, while looser verification executes more actions per full policy inference. SAVE-WAM therefore provides a range of operating points rather than a fixed execution horizon.

\paragraph{Higher efficiency at near-synchronous success rates.}
Compared with synchronous execution, SAVE-WAM supports larger effective action chunks while keeping similar success rates. On RoboTwin-clean, LingBot-VA achieves 92.76\% versus 92.80\% for Sync-16, while increasing the average executed chunk from 16 to 31.55 actions (1.97$\times$). FastWAM achieves 92.52\% versus 92.24\% for Sync-12, with 18.97 versus 12 actions per chunk (1.58$\times$). Completion time also decreases from 97.97 to 91.43s and from 49.79 to 41.02s, respectively. The same trend appears on LIBERO: FastWAM reaches 94.15\% versus 95.20\% for Sync-8, with an average chunk of 15.43 versus 8 and completion time of 7.15 versus 8.16s. These results show that verification enables more executed actions per full policy inference while keeping success close to synchronous execution.

\paragraph{Comparison with adaptive action chunking.}
Table~\ref{tab:robotwin-two-models} compares SAVE-WAM with AAC under asynchronous execution. On RoboTwin-clean, LingBot-VA achieves 92.76\% success with SAVE-WAM versus 85.32\% with AAC, while increasing the average chunk from 27.61 to 31.55 actions, at a higher completion time of 91.43 versus 86.76s. For FastWAM, SAVE-WAM improves both success and efficiency: success rises from 91.72\% to 92.52\%, the average chunk from 16.50 to 18.97, and completion time drops from 43.72 to 41.02s. These results show that observation-based verification provides a stronger success--efficiency trade-off than the evaluated adaptive-chunking baseline, although the time cost varies by backbone.

% ============================LIBERO=========================

\begin{table}[!htbp]
\centering
\caption{LIBERO results with model-specific baselines. SR is success rate, Chunk is the executed chunk size, and Time is the average successful-episode completion time. For LingBot-VA, we report results only on LIBERO-10, as the publicly released checkpoint is trained exclusively on this suite.}
\label{tab:libero-model-separated}
\begingroup
\setlength{\tabcolsep}{3.0pt}
\renewcommand{\arraystretch}{1.05}

\resizebox{\textwidth}{!}{%
\begin{tabular}{@{}l *{12}{r}@{}}
\toprule
\rowcolor{black!5}
\multicolumn{13}{c}{\rule{0pt}{2.4ex}\bfseries\sffamily FastWAM} \\
\midrule
\multirow{2}{*}{Suite}
& \multicolumn{3}{c}{Ours}
& \multicolumn{3}{c}{Sync-8}
& \multicolumn{3}{c}{RTC}
& \multicolumn{3}{c}{AAC} \\
\cmidrule(lr){2-4}
\cmidrule(lr){5-7}
\cmidrule(lr){8-10}
\cmidrule(lr){11-13}
& SR (\%) $\uparrow$ & Chunk $\uparrow$ & Time (s) $\downarrow$
& SR (\%) $\uparrow$ & Chunk $\uparrow$ & Time (s) $\downarrow$
& SR (\%) $\uparrow$ & Chunk $\uparrow$ & Time (s) $\downarrow$
& SR (\%) $\uparrow$ & Chunk $\uparrow$ & Time (s) $\downarrow$ \\
\midrule
libero\_10
& 90.00 & 14.84 & 10.440
& \textbf{92.00} & 8.00 & 12.950
& 75.80 & 15.72 & 9.950
& 76.20 & 16.58 & 13.440 \\

libero\_goal
& 93.00 & 15.58 & 5.820
& \textbf{93.80} & 8.00 & 6.010
& 91.00 & 15.14 & 5.030
& 88.20 & 16.58 & 6.140 \\

libero\_spatial
& 94.60 & 15.22 & 5.960
& \textbf{95.40} & 8.00 & 6.610
& 89.80 & 15.23 & 5.310
& 86.00 & 16.61 & 6.190 \\

libero\_object
& 99.00 & 16.06 & 6.380
& \textbf{99.60} & 8.00 & 7.060
& 93.00 & 15.34 & 5.690
& 89.00 & 16.56 & 7.240 \\
\midrule
Overall
& 94.15 & 15.43 & 7.150
& 95.20 & 8.00 & 8.160
& 87.40 & 15.35 & 6.495
& 84.85 & 16.58 & 8.253 \\
\bottomrule
\end{tabular}%
}

\vspace{7pt}

\resizebox{\textwidth}{!}{%
\begin{tabular}{@{}l *{15}{r}@{}}
\toprule
\rowcolor{black!5}
\multicolumn{16}{c}{\rule{0pt}{2.4ex}\bfseries\sffamily LingBot-VA} \\
\midrule
\multirow{2}{*}{Suite}
& \multicolumn{3}{c}{Ours (Async-16+Verify)}
& \multicolumn{3}{c}{FDM Async-8}
& \multicolumn{3}{c}{FDM Async-16}
& \multicolumn{3}{c}{Sync-8}
& \multicolumn{3}{c}{AAC} \\
\cmidrule(lr){2-4}
\cmidrule(lr){5-7}
\cmidrule(lr){8-10}
\cmidrule(lr){11-13}
\cmidrule(lr){14-16}
& SR (\%) $\uparrow$ & Chunk $\uparrow$ & Time (s) $\downarrow$
& SR (\%) $\uparrow$ & Chunk $\uparrow$ & Time (s) $\downarrow$
& SR (\%) $\uparrow$ & Chunk $\uparrow$ & Time (s) $\downarrow$
& SR (\%) $\uparrow$ & Chunk $\uparrow$ & Time (s) $\downarrow$
& SR (\%) $\uparrow$ & Chunk $\uparrow$ & Time (s) $\downarrow$ \\
\midrule
libero\_10
& 95.80 & 11.45 & 57.159
& 86.60 & 8.00 & 57.293
& 69.40 & 16.00 & 54.873
& \textbf{96.00} & 8.00 & 63.439
& 84.00 & 7.58 & 61.764 \\
\bottomrule
\end{tabular}%
}

\endgroup
\end{table}

{
\subsection{Real-Robot Experiments}
\label{sec:real-robot}

{
In the simulation environment, the iteration time of the simulator dominates, affecting the presentation of end-to-end acceleration. We therefore evaluate SAVE-WAM on two real-robot tasks: \textit{Cube-to-Plate} and \textit{Dynamic Pick}, where the cube is moved during execution to test reactivity. Details are provided in Appendix~\ref{app:real-robot}. As shown in Table~\ref{tab:real-robot}, SAVE-WAM matches Sync-16 success while reducing completion time from $44.1$ to $28.5$,s and from $50.6$ to $33.4$,s, corresponding to $1.55\times$ and $1.51\times$ speedups. Naively increasing the asynchronous horizon to 32 with higher execution frequency is faster but substantially degrades success, whereas verification preserves reliability while improving efficiency.

On the robot, we can raise the action execution frequency while keeping the chunk execution window close to the policy inference time so the two overlap. A longer average executed action horizon supports a higher frequency at the same inference latency, shortening task completion time.
}
}

\begin{table}[!ht]
    \centering
    \scriptsize
    \setlength{\tabcolsep}{2pt}
    \renewcommand{\arraystretch}{0.92}
    \caption{Real-robot results: SR (\%) / time (s).}
    \label{tab:real-robot}
    \begin{adjustbox}{max width=\linewidth}

    \begin{tabular}{lcccc}
        \toprule
        Task & Sync-16 & Async-16 & Async-32 & +SAVE-WAM \\
        \midrule
        Cube-to-Plate
        & 90 / 44.1
        & 80 / 38.7
        & 40 / 23.9
        & \textbf{90 / 28.5} \\
        Dynamic Pick
        & 70 / 50.6
        & 30 / 46.2
        & 10 / 32.8
        & \textbf{70 / 33.4} \\
        \bottomrule
    \end{tabular}
\end{adjustbox}
\end{table}

\subsection{Ablation Study}
\label{sec:ablation}
\begin{table}[!htbp]
\centering
\caption{
Ablation study on two challenging RoboTwin-clean tasks using LingBot-VA
with an Async-64 draft.
\emph{Obs.} denotes the observation used for feedback:
generation-time (\emph{Gen.-time}), execution-time (\emph{Exec.-time}),
or an observation obtained $x$ actions ahead of execution
(\emph{Ahead-$x$}).
Latency and chunk size are averaged over successful episodes.
}
\label{tab:ablation}

\begingroup
\scriptsize
\setlength{\tabcolsep}{2.4pt}
\renewcommand{\arraystretch}{1.12}

\definecolor{basebg}{RGB}{247,247,247}
\definecolor{checkgreen}{RGB}{45,120,85}
\definecolor{crossgray}{RGB}{125,125,125}

\newcommand{\cmark}{\textcolor{checkgreen}{\checkmark}}
\newcommand{\xmark}{\textcolor{crossgray}{$\times$}}

\begin{tabular*}{\textwidth}{@{\extracolsep{\fill}}
    l
    c
    c
    c c
    c
    r r r
    r r r
@{}}
\toprule
\multirow{2}{*}{Variant}
& \multicolumn{1}{c}{Input}
& \multicolumn{1}{c}{Module}
& \multicolumn{2}{c}{Output}
& \multicolumn{1}{c}{Pipeline}
& \multicolumn{3}{c}{\textit{Hanging Mug}}
& \multicolumn{3}{c}{\textit{Place Can Basket}} \\
\cmidrule(lr){2-2}
\cmidrule(lr){3-3}
\cmidrule(lr){4-5}
\cmidrule(lr){6-6}
\cmidrule(lr){7-9}
\cmidrule(lr){10-12}
& Obs.
& Verifier
& Prefix
& Replan
& Overlap
& SR$\uparrow$
& Lat.$\downarrow$
& Chunk$\uparrow$
& SR$\uparrow$
& Lat.$\downarrow$
& Chunk$\uparrow$ \\
\midrule

\rowcolor{basebg}
Async Baseline (64)
& Gen.-time
& --
& Fixed
& \xmark
& --
& 4.00
& 121.281
& 64.00
& 76.00
& 105.451
& 64.00 \\

$+$ Feedback (Sec.~\ref{sec:feedback-output})
& Exec.-time
& Oracle
& Dynamic
& \cmark
& \xmark
& \textbf{40.00}
& 184.074
& \textbf{38.73}
& \textbf{86.00}
& 159.475
& \textbf{37.05} \\

$+$ Low-cost Verify (Sec.~\ref{sec:validity})
& Exec.-time
& One-step
& Dynamic
& \cmark
& \xmark
& 38.00
& 161.565
& 35.61
& 84.00
& 130.840
& 34.65 \\

\rowcolor{resultshade}
$+$ Ahead Verify (Sec.~\ref{sec:ahead-verification})
& Ahead-$x$
& One-step
& Dynamic
& \cmark
& \cmark
& 38.00
& \textbf{138.995}
& 35.03
& 82.00
& \textbf{123.526}
& 34.70 \\

\bottomrule
\end{tabular*}
\endgroup
\end{table}

We ablate SAVE-WAM on \textit{Hanging Mug} and \textit{Place Can Basket} with LingBot-VA and an Async-64 draft. Table~\ref{tab:ablation} studies the three feedback components: \emph{input} controls when the verification observation is acquired, \emph{module} compares oracle regeneration with our one-step verifier, and \emph{output} forms a dynamic accepted prefix through rejection-triggered replanning. Feedback improves Async-64 from $4\%$ to $40\%$ and from $76\%$ to $86\%$ on the two tasks. Replacing the oracle with our verifier reduces success by only $2$ points on both tasks, while ahead verification retains similar success ($38\%/82\%$) and lowers successful-episode latency from $161.565/130.840$\,s to $138.995/123.526$\,s by overlapping verification with execution.

\FloatBarrier
\subsection{Qualitative Example}

\begin{figure}[!htbp]
    \centering

    \includegraphics[width=0.65\linewidth]{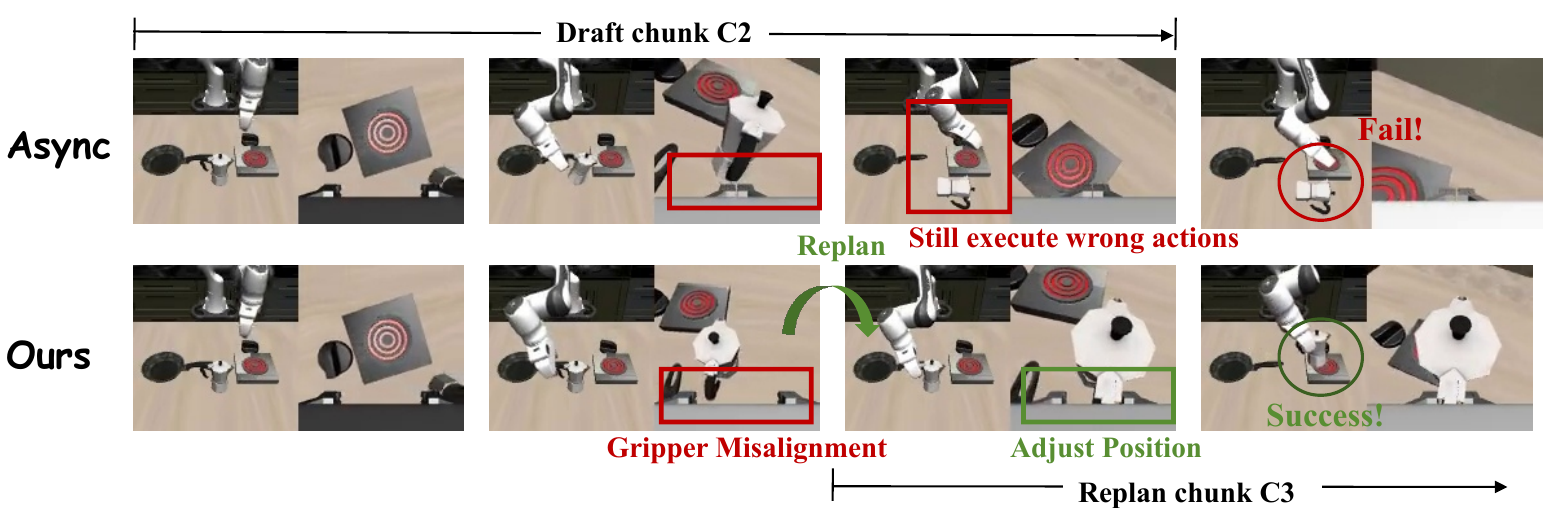}

    \includegraphics[width=0.6\linewidth]{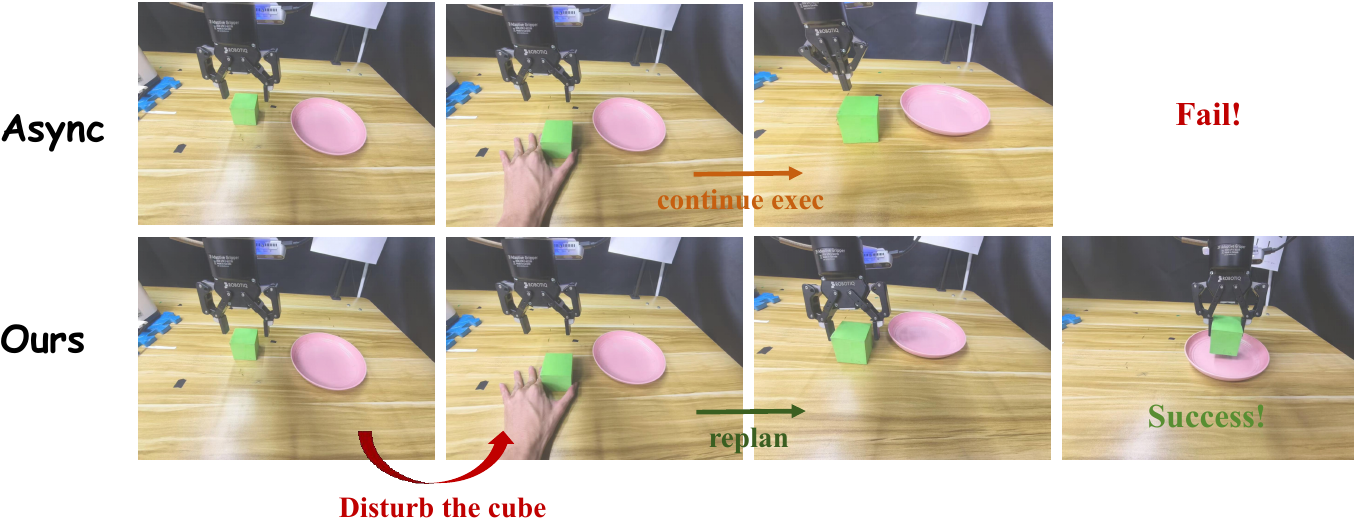}

    \caption{\textbf{Qualitative results in simulation and on the real robot.}
    In simulation, SAVE-WAM detects gripper misalignment and replans instead
    of continuing stale actions. On the real robot, it detects execution-time
    object displacement, rejects the stale draft, and replans from the updated
    observation.}
    \label{fig:qualitative_replan}
\end{figure}

{
Figures~\ref{fig:qualitative_replan} show consistent failure and recovery patterns in simulation and on the real robot. Without updated feedback, asynchronous execution can continue along stale actions after a grasp or scene mismatch. SAVE-WAM detects the deviation, rejects the invalid suffix, and replans from the updated observation to recover the task.
}

\section{Conclusion}
We introduce SAVE-WAM, a training-free feedback mechanism that verifies asynchronous WAM actions with newer observations and triggers replanning when needed. Across simulation and real-robot tasks, SAVE-WAM recovers synchronous-level success while preserving much of latency benefit of asynchronous execution, enabling larger action horizons for up to $1.55\times$ real-robot speedup.

\FloatBarrier

\bibliography{references}
\bibliographystyle{plainnat}

\clearpage

\appendix
\pdfbookmark[0]{Appendix}{appendix}
\section*{Appendix}

\section{One-Step Verification}

\subsection{Proof of the One-Step Validity Bound}
\label{app:one-step-validity-proof}

\begin{proof}
For brevity, define the normalized Euclidean norm
\begin{equation}
    \|X\|_n
    =
    \frac{\|X\|_2}{\sqrt{n}}.
\end{equation}
The oracle discrepancy in Equation~\eqref{eq:oracle-validity} is therefore
\begin{equation}
    d^{\mathrm{oracle}}
    =
    \|A^{\mathrm d}-A^\star\|_n.
\end{equation}

Using the same coupled Gaussian noise $\xi$, define the noisy states associated with the draft and synchronous reference as
\begin{equation}
    Z_\tau^{\mathrm d}
    =
    (1-\tau)A^{\mathrm d}+\tau\xi,
    \qquad
    Z_\tau^\star
    =
    (1-\tau)A^\star+\tau\xi.
\end{equation}
Their corresponding linear-flow target velocities are
\begin{equation}
    v_{\mathrm{target}}^{\mathrm d}
    =
    \xi-A^{\mathrm d},
    \qquad
    v_{\mathrm{target}}^\star
    =
    \xi-A^\star.
\end{equation}
Hence,
\begin{equation}
    d^{\mathrm{oracle}}
    =
    \left\|
        v_{\mathrm{target}}^{\mathrm d}
        -
        v_{\mathrm{target}}^\star
    \right\|_n.
    \label{eq:proof-action-velocity}
\end{equation}

Let
\begin{equation}
    \hat v_\tau^{\mathrm d}
    =
    v_\theta(Z_\tau^{\mathrm d},o^{\mathrm v},\tau)
\end{equation}
be the model vector field used by our one-step verifier. By the triangle inequality,
\begin{align}
d^{\mathrm{oracle}}
\leq\;&
\left\|
    v_{\mathrm{target}}^{\mathrm d}
    -
    \hat v_\tau^{\mathrm d}
\right\|_n
+
\left\|
    \hat v_\tau^{\mathrm d}
    -
    v^\star(Z_\tau^{\mathrm d},o^{\mathrm v},\tau)
\right\|_n
\nonumber\\
&+
\left\|
    v^\star(Z_\tau^{\mathrm d},o^{\mathrm v},\tau)
    -
    v^\star(Z_\tau^\star,o^{\mathrm v},\tau)
\right\|_n
\nonumber\\
&+
\left\|
    v^\star(Z_\tau^\star,o^{\mathrm v},\tau)
    -
    v_{\mathrm{target}}^\star
\right\|_n.
\label{eq:proof-decomposition}
\end{align}

The first term is exactly the one-step verification residual:
\begin{equation}
    \left\|
        v_{\mathrm{target}}^{\mathrm d}
        -
        \hat v_\tau^{\mathrm d}
    \right\|_n
    =
    \sqrt{
        s_\tau(A^{\mathrm d},o^{\mathrm v})
    }.
\end{equation}

By the assumed model approximation bound,
\begin{equation}
    \left\|
        \hat v_\tau^{\mathrm d}
        -
        v^\star(Z_\tau^{\mathrm d},o^{\mathrm v},\tau)
    \right\|_n
    \leq
    \delta_\theta.
\end{equation}
By the $L_x$-Lipschitz assumption,
\begin{align}
&
\left\|
    v^\star(Z_\tau^{\mathrm d},o^{\mathrm v},\tau)
    -
    v^\star(Z_\tau^\star,o^{\mathrm v},\tau)
\right\|_n
\nonumber\\
&\qquad\leq
L_x
\left\|
    Z_\tau^{\mathrm d}
    -
    Z_\tau^\star
\right\|_n
=
L_x(1-\tau)d^{\mathrm{oracle}}.
\end{align}
Finally, the straightness-error assumption gives
\begin{equation}
    \left\|
        v^\star(Z_\tau^\star,o^{\mathrm v},\tau)
        -
        v_{\mathrm{target}}^\star
    \right\|_n
    \leq
    \delta_{\mathrm{str}}.
\end{equation}

Combining these bounds yields
\begin{equation}
    d^{\mathrm{oracle}}
    \leq
    \sqrt{
        s_\tau(A^{\mathrm d},o^{\mathrm v})
    }
    +
    \delta_\theta
    +
    L_x(1-\tau)d^{\mathrm{oracle}}
    +
    \delta_{\mathrm{str}}.
\end{equation}
Therefore, when $L_x(1-\tau)<1$,
\begin{equation}
    d^{\mathrm{oracle}}
    \leq
    \frac{
        \sqrt{s_\tau(A^{\mathrm d},o^{\mathrm v})}
        +
        \delta_\theta
        +
        \delta_{\mathrm{str}}
    }{
        1-L_x(1-\tau)
    },
\end{equation}
which proves Theorem~\ref{thm:one-step-validity}.
\end{proof}

{
\subsection{One-Step Velocity Consistency as a Verification Signal}
\label{sec:one-step-exp}

Our low-cost verifier is motivated by the action flow-matching objective. Given a draft action segment $A^{\mathrm d}$, Gaussian noise $\xi$, and a fixed flow time $\tau$, we construct
\begin{equation}
    Z_\tau^{\mathrm d}
    =
    (1-\tau)A^{\mathrm d}+\tau\xi,
\end{equation}
with the corresponding target velocity
\begin{equation}
    v_{\mathrm{target}}^{\mathrm d}
    =
    \xi-A^{\mathrm d}.
\end{equation}
Using the updated verification observation $o^{\mathrm v}$, a single model evaluation produces
\begin{equation}
    \hat v_\tau^{\mathrm d}
    =
    v_\theta(Z_\tau^{\mathrm d},o^{\mathrm v},\tau).
\end{equation}
We define the one-step consistency score as
\begin{equation}
    s_\tau(A^{\mathrm d},o^{\mathrm v})
    =
    \frac{1}{n}
    \left\|
        \hat v_\tau^{\mathrm d}
        -
        v_{\mathrm{target}}^{\mathrm d}
    \right\|_2^2,
\end{equation}
where $n$ is the number of scalar elements in the action segment.

The clean action implied by this one-step velocity estimate is
\begin{equation}
    \widetilde A_0
    =
    Z_\tau^{\mathrm d}
    -
    \tau\hat v_\tau^{\mathrm d}.
\end{equation}
Since
$Z_\tau^{\mathrm d}
=
A^{\mathrm d}
+
\tau v_{\mathrm{target}}^{\mathrm d}$,
we have
\begin{equation}
    \widetilde A_0-A^{\mathrm d}
    =
    -\tau
    \left(
        \hat v_\tau^{\mathrm d}
        -
        v_{\mathrm{target}}^{\mathrm d}
    \right).
\end{equation}
Therefore,
\begin{equation}
    \frac{
        \left\|
            \widetilde A_0-A^{\mathrm d}
        \right\|_2
    }{\sqrt{n}}
    =
    \tau
    \sqrt{
        s_\tau(A^{\mathrm d},o^{\mathrm v})
    }.
\end{equation}
Thus, the velocity inconsistency is directly proportional to the magnitude of the one-step correction to the draft, motivating $s_\tau$ as a low-cost proxy for disagreement with the fully denoised oracle action.

\begin{figure}[t]
\centering
\includegraphics[width=0.8\linewidth]{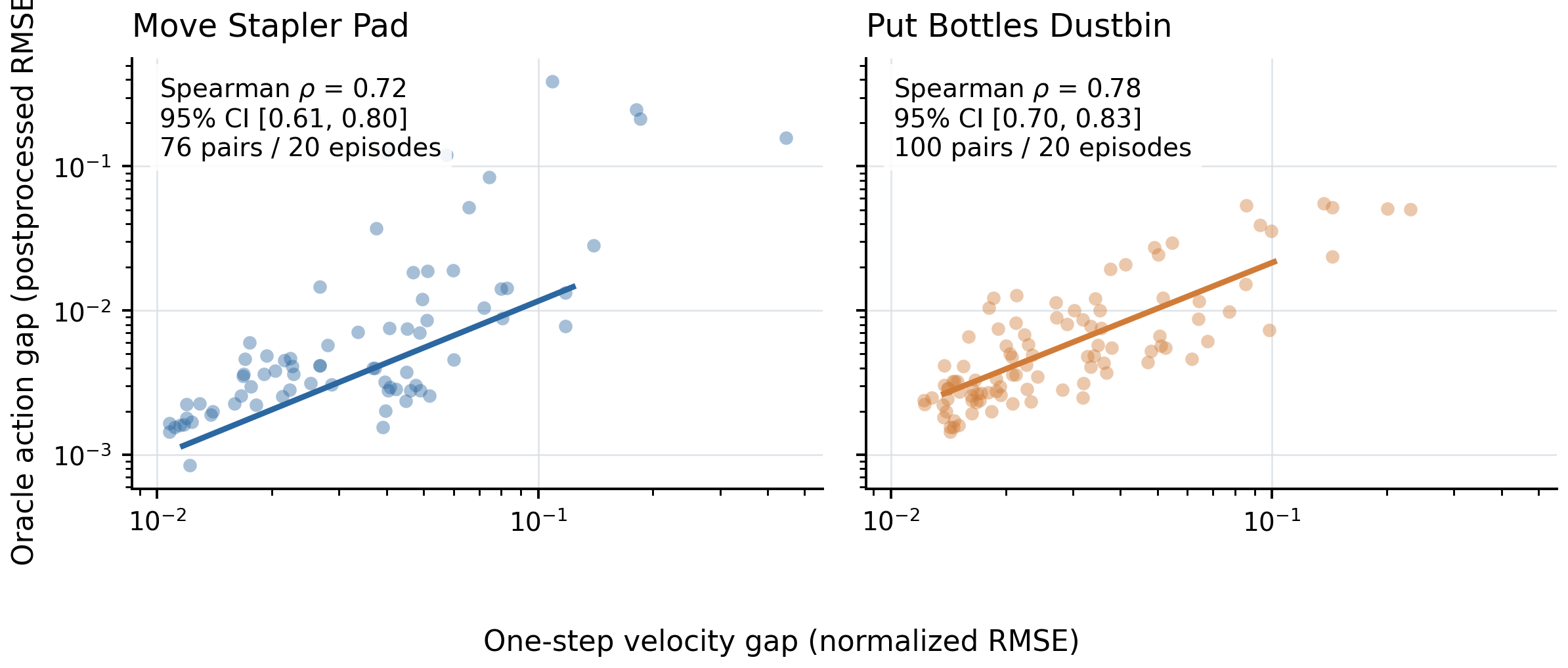}
\caption{\textbf{One-step velocity error vs.\ full-oracle action discrepancy on RoboTwin-Clean.}
Each point represents one paired verification decision; lines show robust log--log trends.}
\label{fig:oracle-correlation}
\end{figure}

Figure~\ref{fig:oracle-correlation} evaluates this relationship over 20 episodes per task. For \textit{Move Stapler Pad}, 76 paired decisions yield a Spearman correlation of $0.719$ (episode-bootstrap 95\% CI: $[0.611,0.799]$) between the one-step score and the full-oracle action discrepancy. For \textit{Put Bottles Dustbin}, 100 paired decisions yield a correlation of $0.779$ ($[0.704,0.832]$). The one-step score also identifies oracle-rejected actions with AUROC values of $0.896$ and $0.900$, respectively. Overall, these results show that a single vector-field evaluation provides an informative proxy for full-oracle disagreement.
}

{
\subsection{Practical Verifier Implementation and Hyperparameters}
\label{app:verifier-details}

This section provides implementation details of the one-step verifier used in our experiments.

\paragraph{Action groups and weighted score.}
All verification is performed in the normalized action space of the corresponding backbone. We partition the action dimensions into three semantic groups: position ($p$), rotation ($r$), and gripper ($g$). For an $h$-step segment, let $\mathcal{I}_k$ denote the action dimensions belonging to group $k\in\{p,r,g\}$ and
\begin{equation}
    n_k=h|\mathcal{I}_k|
\end{equation}
denote the number of scalar residual elements in that group.

Following Equation~\eqref{eq:one-step-residual}, the one-step residual is
\begin{equation}
    r_\tau
    =
    \hat v_\tau^{\mathrm d}
    -
    v_{\mathrm{target}}^{\mathrm d}
    =
    v_\theta(Z_\tau^{\mathrm d},o^{\mathrm v},\tau)
    -
    (\xi-A^{\mathrm d}).
\end{equation}
Let $r_{\tau,k}$ denote the entries of $r_\tau$ belonging to group $k$. We compute
\begin{equation}
    e_k
    =
    \frac{1}{n_k}
    \|r_{\tau,k}\|_2^2,
    \qquad k\in\{p,r,g\}.
\end{equation}

Although the backbone normalizes its action representation, the residual distributions of different semantic groups still have different magnitudes. We therefore use the practical weighted score
\begin{equation}
    S(A^{\mathrm d},o^{\mathrm v})
    =
    0.65e_p+0.20e_r+0.15e_g.
    \label{eq:appendix-weighted-score}
\end{equation}
The weighting balances the contribution of different action groups and avoids overly conservative rejection caused by a single high-magnitude group, thereby allowing a longer average accepted action horizon.

\paragraph{Relation to the theoretical score.}
Theorem~\ref{thm:one-step-validity} is stated using the unweighted one-step score
\begin{equation}
    s_\tau(A^{\mathrm d},o^{\mathrm v})
    =
    \frac{1}{n}\|r_\tau\|_2^2,
    \qquad
    n=\sum_k n_k.
\end{equation}
Let
\begin{equation}
    \alpha_k=\frac{n_k}{n},
    \qquad
    (w_p,w_r,w_g)=(0.65,0.20,0.15).
\end{equation}
Since
\begin{equation}
    s_\tau
    =
    \sum_k \alpha_k e_k
    \leq
    \left(
        \max_k\frac{\alpha_k}{w_k}
    \right)
    \sum_k w_k e_k,
\end{equation}
we obtain
\begin{equation}
    \sqrt{s_\tau}
    \leq
    C_w\sqrt{S},
    \qquad
    C_w
    =
    \sqrt{
        \max_k\frac{\alpha_k}{w_k}
    }.
\end{equation}
Substituting this relation into Theorem~\ref{thm:one-step-validity} yields
\begin{equation}
    d^{\mathrm{oracle}}
    \leq
    \frac{
        C_w\sqrt{S}
        +\delta_\theta
        +\delta_{\mathrm{str}}
    }{
        1-L_x(1-\tau)
    }.
\end{equation}
Thus, the practical weighted score retains the same qualitative connection to the synchronous action discrepancy as the canonical one-step score.

\paragraph{Verification threshold.}
We use one fixed verification threshold for all tasks within each backbone/benchmark configuration rather than tuning it separately for individual tasks. For the configurations reported in the main experiments, we use
\begin{equation}
    \lambda_{\mathrm{ver}}=0.004.
\end{equation}
The threshold controls the reliability--horizon trade-off: a smaller threshold rejects more drafts and produces shorter accepted prefixes, whereas a larger threshold permits longer speculative execution. We select the operating point to recover success close to synchronous execution while retaining longer action horizons.

\paragraph{Noise level.}
We use a fixed verification flow time
\begin{equation}
    \tau=0.3.
\end{equation}
We select $\tau$ by evaluating the agreement between the one-step verifier and the fully denoised oracle under different flow times, following the evaluation protocol in Appendix~\ref{sec:one-step-exp}. The results are shown in Table~\ref{tab:tau-sensitivity}.

\begin{table}[!htbp]
\centering
\caption{Sensitivity of the one-step verifier to the flow time $\tau$.}
\label{tab:tau-sensitivity}
\begin{tabular}{c|ccccc}
\toprule
$\tau$ & 0.1 & 0.2 & \textbf{0.3} & 0.4 & 0.5 \\
\midrule
AUROC & 0.647 & 0.840 & \textbf{0.900} & 0.868 & 0.845 \\
\bottomrule
\end{tabular}
\end{table}

The verifier achieves the highest AUROC at $\tau=0.3$. We therefore fix $\tau=0.3$ throughout the corresponding experiments without task-specific tuning.
}

\FloatBarrier

\section{Asynchronous Execution and Timing}

\subsection{Comparison of our method with Sync and Async}
\begin{figure}[t]
    \centering
    \includegraphics[width=\linewidth]{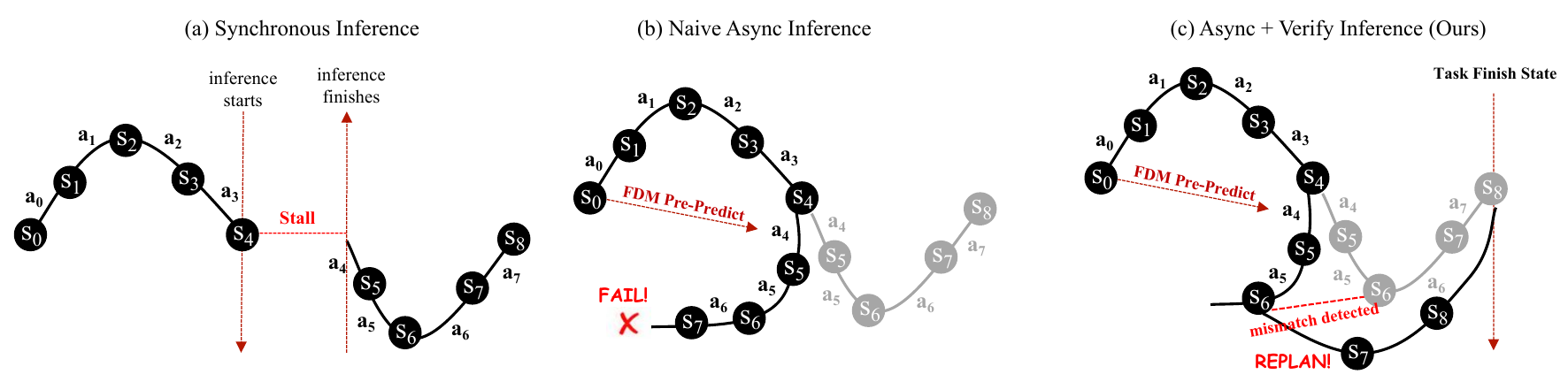}
    \caption{
    \textbf{Overview of feedback-gated asynchronous execution.}
    Synchronous inference provides fresh feedback but suffers from inference
    stalling.
    Naive asynchronous inference improves throughput by generating future actions
    ahead of time, but executes them under stale observations.
    SAVE-WAM closes this feedback gap by verifying speculative actions and
    selectively accepting valid prefixes while replanning upon mismatch.
    }
    \label{fig:overview}
\end{figure}

Figure~\ref{fig:overview} compares synchronous inference, naive asynchronous inference, and SAVE-WAM. Synchronous inference uses the latest observation before generating each action chunk, providing fresh feedback but introducing inference stalls. Asynchronous inference overlaps generation with execution to reduce such stalls, but its actions are generated from older observations and therefore lack feedback from ongoing execution. SAVE-WAM retains asynchronous generation while introducing post-generation feedback: draft actions are verified using newer observations, accepted when they remain consistent, and replanned when a mismatch is detected. This provides a better balance between continuous execution and timely feedback.

{
\subsection{Temporal Alignment for Speculative Ahead Verification}
\label{app:ahead-alignment}

Ahead verification uses an observation acquired before the target segment is executed. Therefore, the newly predicted policy trajectory and the original speculative draft have different temporal origins. This section describes how we align them.

Consider a draft
\begin{equation}
    \mathbf{A}^{\mathrm d}
    =
    [A_1^{\mathrm d},\ldots,A_M^{\mathrm d}],
\end{equation}
where each segment contains $h$ actions. Suppose $A_j^{\mathrm d}$ has already been committed for execution and the next segment $A_{j+1}^{\mathrm d}$ is being verified using a newer observation $o_j^{\mathrm v}$.

A policy trajectory generated from $o_j^{\mathrm v}$ starts from the current execution state. Its first $h$ actions therefore correspond to the already committed interval represented by $A_j^{\mathrm d}$, while the following $h$ actions correspond temporally to the target draft segment $A_{j+1}^{\mathrm d}$.

For clarity, define the block extraction operator
\begin{equation}
    \operatorname{Block}_h(Y,\ell)
    =
    Y[\ell h:(\ell+1)h-1],
\end{equation}
where $\ell=0$ denotes the first $h$ actions of sequence $Y$.
The oracle reference for the next speculative segment is therefore
\begin{equation}
    A^\star_{j+1}
    =
    \operatorname{Block}_h
    \left(
        \pi(o_j^{\mathrm v};\xi),
        1
    \right),
    \label{eq:aligned-oracle-block}
\end{equation}
rather than
$\operatorname{Block}_h(\pi(o_j^{\mathrm v};\xi),0)$.

For example, let $h=8$. At execution step $16$, actions $16$--$23$ are already committed, while actions $24$--$31$ are awaiting verification. The updated observation $o_{16}$ is used to regenerate a reference trajectory. Relative to $o_{16}$, newly generated actions $0$--$7$ correspond to execution steps $16$--$23$, whereas actions $8$--$15$ correspond to steps $24$--$31$. Hence, actions $8$--$15$ form the correct reference for verifying the draft actions $24$--$31$.

The same temporal alignment is used by our one-step verifier. The committed interval is retained as temporal context for the backbone, but it is never reconsidered for execution and does not contribute to the verification residual. Only the aligned target segment $A_{j+1}^{\mathrm d}$ is scored.

For backbones with explicit action-prefix conditioning, such as RTC-based asynchronous execution, the committed actions are supplied through the native prefix-conditioning interface. For future-imagination-based asynchronous execution, the corresponding committed interval is retained in the backbone's action context. These wrappers only differ in how the committed context is represented; the target segment and verification rule are unchanged.

\begin{algorithm}[t]
\caption{Speculative Ahead Verification}
\label{alg:ahead-verification}
\begin{algorithmic}[1]
\REQUIRE Policy $\pi$, verifier $\mathcal{G}$, proposal horizon $H$, segment length $h$
\STATE Observe $o^{\mathrm{gen}}$
\STATE Generate draft $\mathbf{A}^{\mathrm d}\leftarrow\pi(o^{\mathrm{gen}})$
\STATE Partition $\mathbf{A}^{\mathrm d}$ into
$[A_1^{\mathrm d},\ldots,A_M^{\mathrm d}]$, where $M=H/h$
\STATE $j\leftarrow1$
\WHILE{$j\leq M$}
    \STATE Commit $A_j^{\mathrm d}$ for execution
    \IF{$j<M$}
        \STATE Acquire a newer observation $o_j^{\mathrm v}$
        \STATE Retain $A_j^{\mathrm d}$ as committed temporal context
        \STATE In parallel with executing $A_j^{\mathrm d}$,
        compute
        \[
            g_{j+1}
            \leftarrow
            \mathcal{G}
            (
                A_{j+1}^{\mathrm d},
                o_j^{\mathrm v};
                A_j^{\mathrm d}
            )
        \]
        \IF{$g_{j+1}=1$}
            \STATE $j\leftarrow j+1$
        \ELSE
            \STATE Discard
            $[A_{j+1}^{\mathrm d},\ldots,A_M^{\mathrm d}]$
            \STATE Acquire the latest observation $o^{\mathrm{new}}$
            \STATE Replan
            $\mathbf{A}^{\mathrm d}\leftarrow\pi(o^{\mathrm{new}})$
            \STATE Repartition the new draft and set $j\leftarrow1$
        \ENDIF
    \ELSE
        \STATE Acquire the latest observation and generate the next draft
        \STATE $j\leftarrow1$
    \ENDIF
\ENDWHILE
\end{algorithmic}
\end{algorithm}

Because verification is executed concurrently with the current committed segment, its latency can be hidden whenever
$T_{\mathrm{ver}}\leq h\Delta t$.
Thus, ahead verification uses a substantially newer observation than the original generation-time observation while avoiding the execution stall of verification immediately at the segment boundary.}

\subsection{Verification--Execution Pipelining}
\label{app:verify_exec_overlap}

Ahead verification is designed to overlap verification with execution of the
current action segment, thereby minimizing the additional waiting exposed at
the boundary before the next segment can be executed.

\begin{figure}[t]
    \centering
    \includegraphics[width=\linewidth]{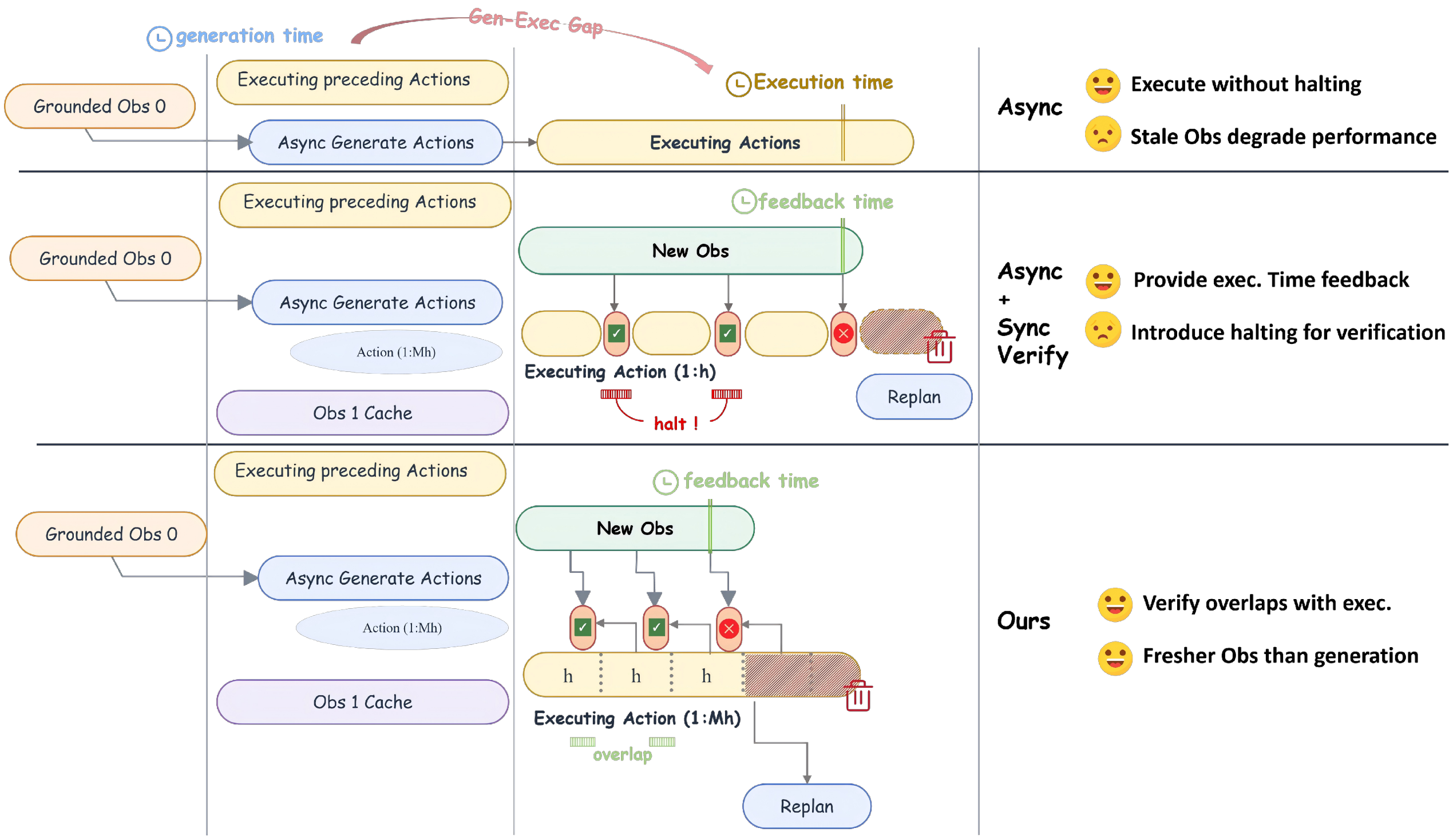}
    \caption{
    \textbf{Verification timing under asynchronous execution.}
    \textit{Top:} Standard asynchronous inference generates actions from an
    earlier observation while preceding actions are still being executed,
    creating a generation--execution gap without post-generation feedback.
    \textit{Middle:} Execution-time synchronous verification uses a fresh
    observation immediately before each segment, but places verification on the
    execution critical path and may introduce stalls.
    \textit{Bottom:} SAVE-WAM moves verification ahead and overlaps it with
    execution of the current segment. Valid segments proceed continuously,
    whereas a detected mismatch rejects the remaining speculative suffix and
    triggers replanning.
    }
    \label{fig:feedback-timing}
\end{figure}

Figure~\ref{fig:feedback-timing} contrasts these timing schemes.
In standard asynchronous execution, the next action chunk is generated before
the preceding actions finish, so the generation observation becomes stale
relative to the state encountered at execution time. The generated actions are
then executed without being reconsidered using newer observations.
A direct alternative is to verify each segment immediately before execution
using the latest observation. Although this provides the freshest feedback, the
verification latency lies on the critical path and can interrupt streaming
execution.

SAVE-WAM instead exploits the substantially lower cost of verification relative
to full action generation and performs it ahead of execution. While the robot
executes the current $h$-step segment, a more recent observation is used to
verify the upcoming speculative segment. If verification succeeds, the next
segment is already available at the execution boundary; if it fails, the
unexecuted suffix is discarded and replanning is triggered. This pipelined
design therefore trades a small amount of observation freshness for substantially
less exposed verification latency.

To quantify this effect, we analyze archived RoboTwin ahead-verification logs
and measure the residual blocking time at each execution boundary. For a
recorded verification event $i$, we define
\begin{equation}
    W_i =
    t_i^{\mathrm{return}}
    -
    t_i^{\mathrm{collect}},
\end{equation}
where $t_i^{\mathrm{collect}}$ denotes the time at which the client requests the
asynchronous verification result through \texttt{future.result()} after
executing the current segment, and $t_i^{\mathrm{return}}$ denotes the time at
which the call returns. Thus, $W_i$ measures the \emph{exposed boundary wait},
rather than the total verification computation time. For cross-draft events,
this quantity may additionally include waiting for the main-model prefetch
dependency.

\begin{table}[!htbp]
    \centering
    \small
    \setlength{\tabcolsep}{4pt}
    \caption{
        Boundary-wait diagnostics for ahead verification with $k=16$.
        The window is the mean time from asynchronous submission to result
        collection, including concurrent execution and any residual waiting.
        All recorded events are included regardless of task success.
        The final column reports the fraction of events with at most
        $1$\,ms exposed wait and should not be interpreted as a compute-overlap
        ratio.
    }
    \label{tab:verify_exec_boundary_wait}
    \begin{tabular}{lrrrrr}
        \toprule
        Event type
        & Events
        & Window
        & Mean $W$
        & P90 $W$
        & $W \leq 1$\,ms \\
        & & (s) & (ms) & (ms) & (\%) \\
        \midrule
        Intra-draft
        & 37,697 & 5.883 & 0.042 & 0.09 & 99.997 \\
        Cross-draft
        & 17,157 & 5.504 & 2.884 & 0.05 & 99.697 \\
        \midrule
        Pooled
        & 54,854 & 5.765 & 0.931 & 0.07 & 99.903 \\
        \bottomrule
    \end{tabular}
\end{table}

As shown in Table~\ref{tab:verify_exec_boundary_wait}, 54,801 of the 54,854
recorded events incur at most $1$\,ms of exposed waiting at the execution
boundary. The pooled mean wait is only $0.931$\,ms, with a 90th percentile of
$0.07$\,ms. The larger mean for cross-draft events, despite their small
90th-percentile value, is caused by a small tail of longer waits associated with
cross-draft dependencies. Overall, these measurements show that verification is
typically completed within the concurrent execution window, leaving negligible
additional blocking on the execution path.

{
\subsection{Timing Breakdown and Inference Overlap}
\label{app:timing-breakdown}

We define the inference overlap ratio as the fraction of accumulated policy inference time that does not block action execution:
\begin{equation}
    R_{\mathrm{overlap}}
    =
    \frac{
        T_{\mathrm{infer}}-T_{\mathrm{blocking}}
    }{
        T_{\mathrm{infer}}
    },
\end{equation}
where $T_{\mathrm{infer}}$ is the accumulated policy inference time and $T_{\mathrm{blocking}}$ is the exposed waiting time caused mainly by verification-triggered synchronous replanning.

For LingBot-VA on RoboTwin-clean, we measure
$T_{\mathrm{infer}}=30.99$\,s and
$T_{\mathrm{blocking}}=8.12$\,s, giving
\begin{equation}
    R_{\mathrm{overlap}}
    =
    \frac{30.99-8.12}{30.99}
    =
    73.8\%.
\end{equation}
Thus, although verification and replanning introduce additional end-to-end latency, most policy inference remains overlapped with action execution. The remaining increase in completion time mainly comes from synchronous replanning after rejected verification, together with the underlying action-execution cost.
}

\FloatBarrier

\section{Supplementary Experiments}

{
\subsection{Additional Baseline Comparisons}
\label{app:additional-baselines}

We further compare SAVE-WAM with additional adaptive execution, uncertainty estimation, and action-correction baselines on the full LIBERO benchmark using FastWAM. All methods are evaluated under the same task protocol, and we report success rate (SR) together with average successful-episode completion time.

\paragraph{EMA and BID.}
EMA applies temporal ensembling by exponentially averaging overlapping actions from consecutive predictions, providing smoother action execution without additional policy sampling. BID instead introduces additional candidate generation and selects actions according to backward coherence with previous decisions and forward contrast between policy predictions. Compared with these methods, SAVE-WAM explicitly uses updated execution-time observations to verify whether pending asynchronous actions remain valid.

\begin{table}[!htbp]
\centering
\small
\caption{Comparison with EMA and BID on LIBERO using FastWAM. Each entry reports SR (\%) / completion time (s).}
\label{tab:ema-bid}
\begin{tabular}{lccc}
\toprule
Suite & EMA & BID & SAVE-WAM \\
\midrule
LIBERO-10
& 78.00 / 9.79
& 87.20 / 12.10
& \textbf{90.00} / 10.44 \\
LIBERO-Goal
& 88.40 / \textbf{4.80}
& 92.60 / 8.66
& \textbf{93.00} / 5.82 \\
LIBERO-Spatial
& 90.00 / 6.05
& 93.80 / 8.73
& \textbf{94.60} / \textbf{5.96} \\
LIBERO-Object
& 94.00 / \textbf{5.49}
& 98.00 / 9.43
& \textbf{99.00} / 6.38 \\
\midrule
Overall
& 87.60 / \textbf{6.34}
& 92.90 / 9.48
& \textbf{94.15} / 7.15 \\
\bottomrule
\end{tabular}
\end{table}

As shown in Table~\ref{tab:ema-bid}, SAVE-WAM achieves the highest success rate on all four suites and improves the overall success rate from $87.60\%$ with EMA and $92.90\%$ with BID to $94.15\%$. EMA introduces less latency but provides substantially smaller reliability gains, while BID requires additional computation and remains slower than SAVE-WAM overall.

\paragraph{AutoHorizon and VLA-Corrector.}
We additionally compare against two closely related approaches. AutoHorizon estimates action uncertainty from the denoising process and adapts the execution horizon accordingly, while VLA-Corrector introduces a learned correction mechanism to revise predicted actions. In contrast, SAVE-WAM is training-free and operates after action generation, allowing it to directly augment asynchronous execution without retraining the underlying policy.

\begin{table}[!htbp]
\centering
\small
\caption{Comparison with related uncertainty- and correction-based methods on LIBERO using FastWAM. Each entry reports SR (\%) / completion time (s).}
\label{tab:related-feedback-baselines}
\begin{tabular}{lccc}
\toprule
Suite & AutoHorizon & VLA-Corrector & SAVE-WAM \\
\midrule
LIBERO-10
& \textbf{91.20} / 12.19
& 88.00 / 18.86
& 90.00 / \textbf{10.44} \\
LIBERO-Goal
& \textbf{94.40} / \textbf{5.23}
& 92.00 / 9.51
& 93.00 / 5.82 \\
LIBERO-Spatial
& 91.80 / \textbf{5.81}
& 93.40 / 9.78
& \textbf{94.60} / 5.96 \\
LIBERO-Object
& 97.40 / 6.42
& 95.40 / 11.15
& \textbf{99.00} / \textbf{6.38} \\
\midrule
Overall
& 93.70 / 7.37
& 92.20 / 13.06
& \textbf{94.15} / \textbf{7.15} \\
\bottomrule
\end{tabular}
\end{table}

As shown in Table~\ref{tab:related-feedback-baselines}, SAVE-WAM achieves the highest overall success rate ($94.15\%$) and the lowest overall completion time ($7.15$\,s). Unlike VLA-Corrector, SAVE-WAM requires no additional training, and unlike the original AutoHorizon formulation, it is designed explicitly for asynchronous action execution.
}

{
\subsection{Analysis of Verification Rejections}
\label{app:rejection-analysis}

We further analyze where verification rejections occur within a speculative draft. We normalize each verification position by the full draft horizon and divide it into three stages: early ($0$--$25\%$), middle ($25$--$75\%$), and late ($75$--$100\%$). For each stage, we report the fraction of verification decisions that are rejected within that positional range.

\begin{table}[!htbp]
\centering
\small
\caption{Rejection rates (\%) at different positions of the speculative draft under different verification thresholds.}
\label{tab:rejection-position}
\begin{tabular}{ccccc}
\toprule
Threshold & Overall & Early & Middle & Late \\
 & rejection & ($0$--$25\%$) & ($25$--$75\%$) & ($75$--$100\%$) \\
\midrule
0.004 & 40.50 & 8.20 & 38.50 & 76.80 \\
0.006 & 34.98 & 7.70 & 34.20 & 63.80 \\
0.008 & 23.03 & 5.20 & 22.70 & 41.50 \\
0.010 & 15.95 & 2.90 & 15.80 & 29.30 \\
\bottomrule
\end{tabular}
\end{table}

As shown in Table~\ref{tab:rejection-position}, rejections consistently become more frequent toward the end of a speculative draft. At the default threshold of $0.004$, the rejection rate increases from only $8.2\%$ in the early stage to $38.5\%$ in the middle and $76.8\%$ in the late stage. The same trend holds across all tested thresholds. Increasing the threshold reduces rejection rates globally, as expected, but does not change this positional pattern.

These results are consistent with the generation--execution discrepancy targeted by SAVE-WAM: actions farther from the generation-time observation are more likely to become inconsistent with the updated execution state. Rather than uniformly shortening every action chunk, verification therefore preserves early draft actions that remain reliable while selectively rejecting later actions when the accumulated discrepancy becomes large.
}

{
\subsection{Real-Robot Experimental Setup}
\label{app:real-robot}

\paragraph{System setup.}
Our real-robot evaluation is implemented based on the RLinf framework with one Franka arm. We use the same policy interface for synchronous and asynchronous execution, and only modify the action scheduling and verification logic for different inference modes. We collect 20 real-robot trajectories for the experimental setup and evaluation. All compared methods use the same observation and action interfaces to ensure a consistent comparison.

\paragraph{Tasks.}
We evaluate two manipulation settings with different levels of execution-time uncertainty.

\textit{Cube-to-Plate} is a standard pick-and-place task. The robot is required to locate and grasp a cube and place it onto the target plate. No external disturbance is introduced during execution. This setting mainly evaluates whether a larger asynchronous action horizon can improve real-world execution efficiency without sacrificing task success.

\textit{Dynamic Pick} introduces an execution-time disturbance to evaluate reactivity. After policy execution has started, the cube is moved from its original position. Consequently, actions generated before the disturbance may no longer be valid under the updated scene. Successful execution therefore requires the policy to respond to the new observation rather than continuing the stale speculative trajectory.

\paragraph{Compared execution modes.}
We compare four settings:
\begin{itemize}
    \item \textbf{Sync-16}: synchronous policy inference with an action horizon of 16;
    \item \textbf{Async-16}: asynchronous execution with a 16-action proposal horizon;
    \item \textbf{Async-32}: asynchronous execution with an enlarged 32-action proposal horizon and higher execution frequency;
    \item \textbf{Async-32 + SAVE-WAM}: the same 32-action asynchronous proposal augmented with our execution-time verification and replanning mechanism.
\end{itemize}

This comparison isolates the effect of increasing the action horizon and whether SAVE-WAM can retain the resulting efficiency gain without the reliability degradation of naive asynchronous execution.

\paragraph{Metrics and protocol.}
We report task success rate and end-to-end task completion time. Success rate is evaluated over 10 trials for each reported method--task configuration. Completion time is averaged over successful trials and measures the wall-clock time from the start of policy execution until successful task completion. Failed trials are excluded from completion-time averaging because they do not have a meaningful task-completion timestamp.

For \textit{Dynamic Pick}, the object is deliberately displaced during policy execution so that the pending asynchronous actions are generated from a stale scene state. SAVE-WAM continues to use updated observations during execution; once the pending draft fails verification, its unexecuted suffix is discarded and the policy replans from the updated observation.

\paragraph{Effect of action horizon.}
The real-robot results reveal a clear reliability--latency trade-off. Increasing the asynchronous horizon from 16 to 32 substantially reduces completion time, but also reduces success from $8/10$ to $4/10$ on \textit{Cube-to-Plate} and from $3/10$ to $1/10$ on \textit{Dynamic Pick}. SAVE-WAM retains the 32-action proposal horizon while recovering success to $9/10$ and $7/10$, respectively. Compared with Sync-16 at the same success rate, this corresponds to $1.55\times$ and $1.51\times$ end-to-end speedups on the two tasks.
}

\subsection{Full Task Results on RoboTwin 2.0}
\label{app:full_task_results}

This section reports the full per-task results on RoboTwin 2.0 in Table~\ref{tab:robotwin-clean-random} and Table~\ref{tab:fastwam-robotwin-clean-random}, complementing
the representative examples shown in the main paper.

\begingroup
\scriptsize
\setlength{\tabcolsep}{1.45pt}
\setlength{\LTcapwidth}{\linewidth}
\setlength{\LTleft}{-0.45in plus 1fill}
\setlength{\LTright}{-0.45in plus 1fill}
\newcommand{\resulttablehead}{%
\toprule
\multirow{2}{*}{Task} & \multirow{2}{*}{Setting} & \multicolumn{3}{c}{Ours} & \multicolumn{3}{c}{FDM Async-32} & \multicolumn{3}{c}{FDM Async-64} & \multicolumn{3}{c}{Sync-16} & \multicolumn{3}{c}{AAC} \\
\cmidrule(lr){3-5}\cmidrule(lr){6-8}\cmidrule(lr){9-11}\cmidrule(lr){12-14}\cmidrule(lr){15-17}
 &  & SR $\uparrow$ & Chunk & Time (s) & SR $\uparrow$ & Chunk & Time (s) & SR $\uparrow$ & Chunk & Time (s) & SR $\uparrow$ & Chunk & Time (s) & SR $\uparrow$ & Chunk & Time (s) \\
\midrule
}
\begin{longtable}{@{}l c *{5}{r r r}@{}}
\caption{Per-task results of LingBot-VA on RoboTwin. E and H denote the clean (easy) and random (hard) settings, respectively. Time is the successful episode average completion time of each task. }\label{tab:robotwin-clean-random}\\
\resulttablehead
\endfirsthead
\multicolumn{1}{@{}l}{\tablename~\thetable\ (continued)} \\
\resulttablehead
\endhead
\multirow{2}{*}{\textit{Adjust Bottle}} & E & 98 & 29.98 & \multirow{2}{*}{56.873} & 96 & 32 & \multirow{2}{*}{49.510} & 94 & 64 & \multirow{2}{*}{46.894} & \textbf{100} & 16 & \multirow{2}{*}{59.762} & 94 & 25.89 & \multirow{2}{*}{52.215} \\*
 & H & 98 & 27.77 &  & 96 & 32 &  & \textbf{100} & 64 &  & 98 & 16 &  & 94 & 26.59 &  \\
\multirow{2}{*}{\textit{Beat Block Hammer}} & E & \textbf{98} & 28.22 & \multirow{2}{*}{53.645} & 90 & 32 & \multirow{2}{*}{47.031} & 92 & 64 & \multirow{2}{*}{45.052} & 96 & 16 & \multirow{2}{*}{55.941} & 80 & 25.38 & \multirow{2}{*}{49.135} \\*
 & H & \textbf{100} & 27.36 &  & 76 & 32 &  & 80 & 64 &  & 96 & 16 &  & 74 & 26.43 &  \\
\multirow{2}{*}{\textit{Blocks Ranking RGB}} & E & 96 & 36.05 & \multirow{2}{*}{181.772} & 90 & 32 & \multirow{2}{*}{158.996} & 86 & 64 & \multirow{2}{*}{154.400} & \textbf{98} & 16 & \multirow{2}{*}{198.278} & 94 & 31.27 & \multirow{2}{*}{170.806} \\*
 & H & \textbf{100} & 29.09 &  & 92 & 32 &  & 88 & 64 &  & 94 & 16 &  & 96 & 31.63 &  \\
\multirow{2}{*}{\textit{Blocks Ranking Size}} & E & \textbf{96} & 35.66 & \multirow{2}{*}{186.307} & 86 & 32 & \multirow{2}{*}{160.453} & 78 & 64 & \multirow{2}{*}{148.515} & 94 & 16 & \multirow{2}{*}{209.906} & 82 & 31.18 & \multirow{2}{*}{165.439} \\*
 & H & \textbf{94} & 29.05 &  & 82 & 32 &  & 60 & 64 &  & 92 & 16 &  & 68 & 31.01 &  \\
\multirow{2}{*}{\textit{Click Alarmclock}} & E & \textbf{100} & 28.50 & \multirow{2}{*}{31.215} & \textbf{100} & 32 & \multirow{2}{*}{28.963} & \textbf{100} & 64 & \multirow{2}{*}{26.414} & \textbf{100} & 16 & \multirow{2}{*}{31.704} & \textbf{100} & 23.59 & \multirow{2}{*}{31.662} \\*
 & H & 98 & 26.06 &  & 90 & 32 &  & \textbf{100} & 64 &  & \textbf{100} & 16 &  & \textbf{100} & 24.24 &  \\
\multirow{2}{*}{\textit{Click Bell}} & E & \textbf{100} & 23.71 & \multirow{2}{*}{30.055} & \textbf{100} & 32 & \multirow{2}{*}{26.923} & \textbf{100} & 64 & \multirow{2}{*}{25.088} & \textbf{100} & 16 & \multirow{2}{*}{30.929} & \textbf{100} & 27.32 & \multirow{2}{*}{27.192} \\*
 & H & \textbf{100} & 26.16 &  & \textbf{100} & 32 &  & \textbf{100} & 64 &  & \textbf{100} & 16 &  & \textbf{100} & 25.02 &  \\
\multirow{2}{*}{\textit{Dump Bin Bigbin}} & E & \textbf{100} & 32.28 & \multirow{2}{*}{84.250} & 92 & 32 & \multirow{2}{*}{79.491} & 88 & 64 & \multirow{2}{*}{65.127} & 92 & 16 & \multirow{2}{*}{89.037} & 94 & 29.02 & \multirow{2}{*}{73.592} \\*
 & H & 98 & 28.23 &  & 96 & 32 &  & 94 & 64 &  & \textbf{100} & 16 &  & 98 & 28.93 &  \\
\multirow{2}{*}{\textit{Grab Roller}} & E & \textbf{100} & 31.67 & \multirow{2}{*}{42.559} & \textbf{100} & 32 & \multirow{2}{*}{39.808} & \textbf{100} & 64 & \multirow{2}{*}{38.024} & \textbf{100} & 16 & \multirow{2}{*}{47.370} & \textbf{100} & 23.98 & \multirow{2}{*}{42.021} \\*
 & H & \textbf{100} & 27.86 &  & 96 & 32 &  & \textbf{100} & 64 &  & \textbf{100} & 16 &  & \textbf{100} & 24.60 &  \\
\multirow{2}{*}{\textit{Handover Block}} & E & 94 & 32.75 & \multirow{2}{*}{122.557} & 64 & 32 & \multirow{2}{*}{102.621} & 60 & 64 & \multirow{2}{*}{98.670} & \textbf{100} & 16 & \multirow{2}{*}{128.638} & 82 & 28.33 & \multirow{2}{*}{109.414} \\*
 & H & 72 & 29.42 &  & 66 & 32 &  & 62 & 64 &  & \textbf{94} & 16 &  & 62 & 28.45 &  \\
\multirow{2}{*}{\textit{Handover Mic}} & E & 96 & 33.64 & \multirow{2}{*}{98.739} & \textbf{100} & 32 & \multirow{2}{*}{86.346} & 92 & 64 & \multirow{2}{*}{81.747} & 98 & 16 & \multirow{2}{*}{105.132} & 78 & 30.72 & \multirow{2}{*}{86.541} \\*
 & H & 96 & 28.62 &  & 96 & 32 &  & 94 & 64 &  & \textbf{98} & 16 &  & 88 & 30.40 &  \\
\multirow{2}{*}{\textit{Hanging Mug}} & E & \textbf{38} & 35.03 & \multirow{2}{*}{138.995} & 26 & 32 & \multirow{2}{*}{128.688} & 4 & 64 & \multirow{2}{*}{121.281} & 30 & 16 & \multirow{2}{*}{156.605} & 20 & 30.65 & \multirow{2}{*}{135.278} \\*
 & H & 42 & 28.03 &  & 30 & 32 &  & 2 & 64 &  & \textbf{46} & 16 &  & 16 & 30.83 &  \\
\multirow{2}{*}{\textit{Lift Pot}} & E & 98 & 28.58 & \multirow{2}{*}{58.653} & 92 & 32 & \multirow{2}{*}{54.038} & \textbf{100} & 64 & \multirow{2}{*}{51.138} & \textbf{100} & 16 & \multirow{2}{*}{61.929} & 70 & 26.65 & \multirow{2}{*}{55.131} \\*
 & H & \textbf{100} & 27.62 &  & 96 & 32 &  & \textbf{100} & 64 &  & 96 & 16 &  & 86 & 27.24 &  \\
\multirow{2}{*}{\textit{Move Can Pot}} & E & 96 & 29.88 & \multirow{2}{*}{68.693} & 84 & 32 & \multirow{2}{*}{60.265} & 88 & 64 & \multirow{2}{*}{56.622} & \textbf{98} & 16 & \multirow{2}{*}{74.735} & 94 & 26.12 & \multirow{2}{*}{63.904} \\*
 & H & 90 & 28.14 &  & 86 & 32 &  & \textbf{98} & 64 &  & 82 & 16 &  & 80 & 26.45 &  \\
\multirow{2}{*}{\textit{Move Pillbottle Pad}} & E & \textbf{100} & 29.13 & \multirow{2}{*}{62.382} & 92 & 32 & \multirow{2}{*}{53.415} & 90 & 64 & \multirow{2}{*}{50.997} & \textbf{100} & 16 & \multirow{2}{*}{68.700} & 94 & 27.22 & \multirow{2}{*}{56.404} \\*
 & H & 94 & 28.43 &  & 90 & 32 &  & 88 & 64 &  & \textbf{96} & 16 &  & 84 & 26.48 &  \\
\multirow{2}{*}{\textit{Move Playingcard Away}} & E & \textbf{100} & 27.06 & \multirow{2}{*}{53.437} & 98 & 32 & \multirow{2}{*}{43.763} & 98 & 64 & \multirow{2}{*}{41.430} & \textbf{100} & 16 & \multirow{2}{*}{56.544} & 94 & 27.56 & \multirow{2}{*}{48.122} \\*
 & H & \textbf{100} & 27.05 &  & 98 & 32 &  & 98 & 64 &  & 94 & 16 &  & 94 & 29.17 &  \\
\multirow{2}{*}{\textit{Move Stapler Pad}} & E & 72 & 29.91 & \multirow{2}{*}{80.236} & 38 & 32 & \multirow{2}{*}{61.609} & 32 & 64 & \multirow{2}{*}{62.797} & \textbf{76} & 16 & \multirow{2}{*}{80.733} & 50 & 29.33 & \multirow{2}{*}{69.516} \\*
 & H & \textbf{54} & 27.88 &  & 44 & 32 &  & 42 & 64 &  & \textbf{54} & 16 &  & 52 & 27.84 &  \\
\multirow{2}{*}{\textit{Open Laptop}} & E & \textbf{96} & 34.86 & \multirow{2}{*}{72.125} & \textbf{96} & 32 & \multirow{2}{*}{68.166} & 90 & 64 & \multirow{2}{*}{68.016} & 94 & 16 & \multirow{2}{*}{88.180} & \textbf{96} & 24.54 & \multirow{2}{*}{79.815} \\*
 & H & 92 & 30.38 &  & 84 & 32 &  & \textbf{100} & 64 &  & 96 & 16 &  & 96 & 24.29 &  \\
\multirow{2}{*}{\textit{Open Microwave}} & E & 84 & 37.35 & \multirow{2}{*}{184.449} & 60 & 32 & \multirow{2}{*}{173.047} & 56 & 64 & \multirow{2}{*}{168.349} & \textbf{94} & 16 & \multirow{2}{*}{218.494} & 58 & 28.28 & \multirow{2}{*}{274.445} \\*
 & H & 72 & 33.50 &  & 56 & 32 &  & 40 & 64 &  & \textbf{98} & 16 &  & 50 & 29.28 &  \\
\multirow{2}{*}{\textit{Pick Diverse Bottles}} & E & \textbf{98} & 27.62 & \multirow{2}{*}{55.064} & 80 & 32 & \multirow{2}{*}{47.197} & 82 & 64 & \multirow{2}{*}{44.490} & 90 & 16 & \multirow{2}{*}{56.574} & 86 & 22.48 & \multirow{2}{*}{49.867} \\*
 & H & 80 & 27.32 &  & \textbf{82} & 32 &  & 80 & 64 &  & 78 & 16 &  & 80 & 22.36 &  \\
\multirow{2}{*}{\textit{Pick Dual Bottles}} & E & \textbf{100} & 27.82 & \multirow{2}{*}{54.979} & 88 & 32 & \multirow{2}{*}{47.857} & 80 & 64 & \multirow{2}{*}{44.269} & \textbf{100} & 16 & \multirow{2}{*}{56.161} & \textbf{100} & 22.61 & \multirow{2}{*}{50.671} \\*
 & H & \textbf{94} & 27.45 &  & 86 & 32 &  & 90 & 64 &  & 82 & 16 &  & 86 & 21.71 &  \\
\multirow{2}{*}{\textit{Place A2B Left}} & E & 94 & 28.76 & \multirow{2}{*}{64.933} & 88 & 32 & \multirow{2}{*}{55.967} & 88 & 64 & \multirow{2}{*}{53.842} & \textbf{96} & 16 & \multirow{2}{*}{70.661} & 90 & 27.13 & \multirow{2}{*}{63.221} \\*
 & H & \textbf{96} & 28.19 &  & 94 & 32 &  & 92 & 64 &  & \textbf{96} & 16 &  & 92 & 26.87 &  \\
\multirow{2}{*}{\textit{Place A2B Right}} & E & 92 & 31.28 & \multirow{2}{*}{62.502} & \textbf{98} & 32 & \multirow{2}{*}{56.781} & 88 & 64 & \multirow{2}{*}{54.225} & 94 & 16 & \multirow{2}{*}{68.782} & 84 & 26.13 & \multirow{2}{*}{58.280} \\*
 & H & 96 & 28.15 &  & 88 & 32 &  & 82 & 64 &  & \textbf{98} & 16 &  & 96 & 25.70 &  \\
\multirow{2}{*}{\textit{Place Bread Basket}} & E & \textbf{98} & 32.10 & \multirow{2}{*}{99.085} & 86 & 32 & \multirow{2}{*}{85.876} & 88 & 64 & \multirow{2}{*}{81.110} & 92 & 16 & \multirow{2}{*}{104.742} & 92 & 27.96 & \multirow{2}{*}{89.718} \\*
 & H & 94 & 28.08 &  & 86 & 32 &  & 84 & 64 &  & \textbf{98} & 16 &  & 80 & 27.03 &  \\
\multirow{2}{*}{\textit{Place Bread Skillet}} & E & 94 & 28.02 & \multirow{2}{*}{65.645} & \textbf{96} & 32 & \multirow{2}{*}{59.391} & 94 & 64 & \multirow{2}{*}{58.671} & 94 & 16 & \multirow{2}{*}{71.603} & 90 & 26.21 & \multirow{2}{*}{63.851} \\*
 & H & \textbf{92} & 28 &  & 86 & 32 &  & 88 & 64 &  & 86 & 16 &  & 82 & 24.81 &  \\
\multirow{2}{*}{\textit{Place Burger Fries}} & E & 96 & 34.64 & \multirow{2}{*}{115.095} & \textbf{100} & 32 & \multirow{2}{*}{97.388} & 96 & 64 & \multirow{2}{*}{90.321} & 96 & 16 & \multirow{2}{*}{120.452} & 96 & 27.36 & \multirow{2}{*}{98.655} \\*
 & H & 94 & 28.98 &  & \textbf{100} & 32 &  & 96 & 64 &  & \textbf{100} & 16 &  & 94 & 27.19 &  \\
\multirow{2}{*}{\textit{Place Can Basket}} & E & 82 & 34.70 & \multirow{2}{*}{123.526} & 74 & 32 & \multirow{2}{*}{113.822} & 76 & 64 & \multirow{2}{*}{105.451} & \textbf{84} & 16 & \multirow{2}{*}{133.699} & 70 & 30.19 & \multirow{2}{*}{123.936} \\*
 & H & \textbf{88} & 29.02 &  & 80 & 32 &  & 80 & 64 &  & 86 & 16 &  & 80 & 30.34 &  \\
\multirow{2}{*}{\textit{Place Cans Plasticbox}} & E & \textbf{100} & 34.71 & \multirow{2}{*}{116.024} & \textbf{100} & 32 & \multirow{2}{*}{100.785} & 92 & 64 & \multirow{2}{*}{96.310} & \textbf{100} & 16 & \multirow{2}{*}{122.469} & \textbf{100} & 29.12 & \multirow{2}{*}{103.337} \\*
 & H & \textbf{100} & 29.15 &  & \textbf{100} & 32 &  & 90 & 64 &  & \textbf{100} & 16 &  & 98 & 28.73 &  \\
\multirow{2}{*}{\textit{Place Container Plate}} & E & \textbf{100} & 33.41 & \multirow{2}{*}{64.026} & \textbf{100} & 32 & \multirow{2}{*}{55.998} & 98 & 64 & \multirow{2}{*}{53.141} & \textbf{100} & 16 & \multirow{2}{*}{67.778} & \textbf{100} & 27.08 & \multirow{2}{*}{58.056} \\*
 & H & \textbf{98} & 28.79 &  & 94 & 32 &  & \textbf{98} & 64 &  & 96 & 16 &  & 96 & 26.98 &  \\
\multirow{2}{*}{\textit{Place Dual Shoes}} & E & 92 & 30.55 & \multirow{2}{*}{116.915} & 88 & 32 & \multirow{2}{*}{103.244} & 80 & 64 & \multirow{2}{*}{100.766} & \textbf{96} & 16 & \multirow{2}{*}{124.156} & 86 & 28.45 & \multirow{2}{*}{105.115} \\*
 & H & 90 & 28.20 &  & 86 & 32 &  & 86 & 64 &  & \textbf{92} & 16 &  & 84 & 26.20 &  \\
\multirow{2}{*}{\textit{Place Empty Cup}} & E & \textbf{100} & 32.46 & \multirow{2}{*}{69.181} & 96 & 32 & \multirow{2}{*}{61.361} & \textbf{100} & 64 & \multirow{2}{*}{58.775} & \textbf{100} & 16 & \multirow{2}{*}{76.610} & \textbf{100} & 28.78 & \multirow{2}{*}{65.372} \\*
 & H & \textbf{100} & 27.76 &  & \textbf{100} & 32 &  & \textbf{100} & 64 &  & \textbf{100} & 16 &  & 98 & 27.33 &  \\
\multirow{2}{*}{\textit{Place Fan}} & E & \textbf{94} & 28.81 & \multirow{2}{*}{66.664} & 80 & 32 & \multirow{2}{*}{56.908} & 80 & 64 & \multirow{2}{*}{54.146} & 92 & 16 & \multirow{2}{*}{72.645} & 92 & 27.83 & \multirow{2}{*}{61.947} \\*
 & H & 86 & 28.64 &  & 78 & 32 &  & 80 & 64 &  & \textbf{88} & 16 &  & 86 & 26.92 &  \\
\multirow{2}{*}{\textit{Place Mouse Pad}} & E & 84 & 32.31 & \multirow{2}{*}{67.663} & 74 & 32 & \multirow{2}{*}{60.938} & 72 & 64 & \multirow{2}{*}{55.205} & \textbf{94} & 16 & \multirow{2}{*}{71.768} & 82 & 27.04 & \multirow{2}{*}{61.146} \\*
 & H & 88 & 28.34 &  & 76 & 32 &  & 70 & 64 &  & \textbf{96} & 16 &  & 78 & 28.31 &  \\
\multirow{2}{*}{\textit{Place Object Basket}} & E & \textbf{86} & 33.07 & \multirow{2}{*}{121.307} & 76 & 32 & \multirow{2}{*}{105.033} & 76 & 64 & \multirow{2}{*}{103.648} & 80 & 16 & \multirow{2}{*}{128.577} & 76 & 29.53 & \multirow{2}{*}{108.944} \\*
 & H & \textbf{86} & 28.64 &  & 80 & 32 &  & 70 & 64 &  & 78 & 16 &  & 64 & 31.61 &  \\
\multirow{2}{*}{\textit{Place Object Scale}} & E & \textbf{100} & 28.14 & \multirow{2}{*}{63.144} & 96 & 32 & \multirow{2}{*}{56.317} & 92 & 64 & \multirow{2}{*}{54.001} & \textbf{100} & 16 & \multirow{2}{*}{68.638} & 90 & 26.33 & \multirow{2}{*}{63.812} \\*
 & H & \textbf{96} & 28.21 &  & 88 & 32 &  & 88 & 64 &  & 94 & 16 &  & 84 & 25.78 &  \\
\multirow{2}{*}{\textit{Place Object Stand}} & E & 96 & 31.26 & \multirow{2}{*}{62.011} & 96 & 32 & \multirow{2}{*}{54.374} & \textbf{98} & 64 & \multirow{2}{*}{50.893} & \textbf{98} & 16 & \multirow{2}{*}{65.147} & 94 & 28.01 & \multirow{2}{*}{55.379} \\*
 & H & 94 & 28.60 &  & \textbf{98} & 32 &  & \textbf{98} & 64 &  & \textbf{98} & 16 &  & 92 & 27.35 &  \\
\multirow{2}{*}{\textit{Place Phone Stand}} & E & \textbf{94} & 26.04 & \multirow{2}{*}{54.567} & 74 & 32 & \multirow{2}{*}{48.666} & 70 & 64 & \multirow{2}{*}{47.915} & 92 & 16 & \multirow{2}{*}{60.312} & 90 & 30.28 & \multirow{2}{*}{52.258} \\*
 & H & 96 & 26.48 &  & 90 & 32 &  & \textbf{98} & 64 &  & 96 & 16 &  & 90 & 30.43 &  \\
\multirow{2}{*}{\textit{Place Shoe}} & E & 98 & 31.80 & \multirow{2}{*}{72.047} & 94 & 32 & \multirow{2}{*}{60.455} & 94 & 64 & \multirow{2}{*}{57.482} & \textbf{100} & 16 & \multirow{2}{*}{83.048} & 96 & 28.09 & \multirow{2}{*}{62.476} \\*
 & H & 96 & 28.35 &  & 94 & 32 &  & \textbf{100} & 64 &  & 96 & 16 &  & 96 & 27.82 &  \\
\multirow{2}{*}{\textit{Press Stapler}} & E & 86 & 31.49 & \multirow{2}{*}{50.360} & \textbf{88} & 32 & \multirow{2}{*}{49.261} & 86 & 64 & \multirow{2}{*}{51.812} & 82 & 16 & \multirow{2}{*}{54.852} & 86 & 22.77 & \multirow{2}{*}{53.762} \\*
 & H & 80 & 29.33 &  & 80 & 32 &  & \textbf{96} & 64 &  & 84 & 16 &  & 80 & 23.42 &  \\
\multirow{2}{*}{\textit{Put Bottles Dustbin}} & E & 80 & 32.72 & \multirow{2}{*}{275.897} & 62 & 32 & \multirow{2}{*}{235.294} & 46 & 64 & \multirow{2}{*}{233.738} & \textbf{82} & 16 & \multirow{2}{*}{273.918} & 58 & 31.53 & \multirow{2}{*}{248.468} \\*
 & H & 78 & 29.46 &  & 60 & 32 &  & 48 & 64 &  & \textbf{80} & 16 &  & 58 & 31.29 &  \\
\multirow{2}{*}{\textit{Put Object Cabinet}} & E & 74 & 29.81 & \multirow{2}{*}{140.887} & 74 & 32 & \multirow{2}{*}{119.691} & \textbf{78} & 64 & \multirow{2}{*}{112.570} & 76 & 16 & \multirow{2}{*}{144.997} & 58 & 29.68 & \multirow{2}{*}{123.123} \\*
 & H & \textbf{82} & 28.02 &  & 70 & 32 &  & 62 & 64 &  & 80 & 16 &  & 68 & 31.77 &  \\
\multirow{2}{*}{\textit{Rotate QRcode}} & E & \textbf{98} & 33.36 & \multirow{2}{*}{65.569} & 92 & 32 & \multirow{2}{*}{57.623} & \textbf{98} & 64 & \multirow{2}{*}{56.590} & \textbf{98} & 16 & \multirow{2}{*}{70.962} & 96 & 27.28 & \multirow{2}{*}{62.715} \\*
 & H & 90 & 28.33 &  & 82 & 32 &  & 94 & 64 &  & \textbf{96} & 16 &  & 86 & 25.60 &  \\
\multirow{2}{*}{\textit{Scan Object}} & E & \textbf{88} & 32.72 & \multirow{2}{*}{77.713} & 86 & 32 & \multirow{2}{*}{69.916} & 84 & 64 & \multirow{2}{*}{65.545} & \textbf{88} & 16 & \multirow{2}{*}{80.481} & 84 & 25.91 & \multirow{2}{*}{70.836} \\*
 & H & 90 & 29.11 &  & 86 & 32 &  & \textbf{94} & 64 &  & 88 & 16 &  & 68 & 25.21 &  \\
\multirow{2}{*}{\textit{Shake Bottle}} & E & \textbf{100} & 27.02 & \multirow{2}{*}{47.792} & 98 & 32 & \multirow{2}{*}{42.126} & 98 & 64 & \multirow{2}{*}{39.142} & \textbf{100} & 16 & \multirow{2}{*}{49.830} & 96 & 24.36 & \multirow{2}{*}{44.217} \\*
 & H & 92 & 27.77 &  & 94 & 32 &  & 96 & 64 &  & \textbf{98} & 16 &  & 94 & 24.47 &  \\
\multirow{2}{*}{\textit{Shake Bottle Horizontally}} & E & 98 & 32.85 & \multirow{2}{*}{46.919} & 98 & 32 & \multirow{2}{*}{41.995} & \textbf{100} & 64 & \multirow{2}{*}{39.047} & \textbf{100} & 16 & \multirow{2}{*}{50.218} & \textbf{100} & 24.31 & \multirow{2}{*}{43.873} \\*
 & H & 94 & 27.69 &  & 90 & 32 &  & 94 & 64 &  & \textbf{96} & 16 &  & 94 & 24.15 &  \\
\multirow{2}{*}{\textit{Stack Blocks Three}} & E & \textbf{98} & 36.41 & \multirow{2}{*}{184.893} & 60 & 32 & \multirow{2}{*}{157.519} & 54 & 64 & \multirow{2}{*}{149.612} & \textbf{98} & 16 & \multirow{2}{*}{197.193} & 78 & 31.40 & \multirow{2}{*}{167.722} \\*
 & H & 96 & 29.38 &  & 56 & 32 &  & 52 & 64 &  & \textbf{98} & 16 &  & 68 & 30.97 &  \\
\multirow{2}{*}{\textit{Stack Blocks Two}} & E & \textbf{100} & 36.10 & \multirow{2}{*}{121.614} & \textbf{100} & 32 & \multirow{2}{*}{105.551} & \textbf{100} & 64 & \multirow{2}{*}{99.140} & 98 & 16 & \multirow{2}{*}{132.231} & \textbf{100} & 30.99 & \multirow{2}{*}{110.059} \\*
 & H & \textbf{100} & 28.86 &  & 98 & 32 &  & 94 & 64 &  & \textbf{100} & 16 &  & \textbf{100} & 31.05 &  \\
\multirow{2}{*}{\textit{Stack Bowls Three}} & E & 80 & 37.85 & \multirow{2}{*}{206.929} & 82 & 32 & \multirow{2}{*}{185.361} & 68 & 64 & \multirow{2}{*}{182.426} & \textbf{86} & 16 & \multirow{2}{*}{216.308} & 76 & 30.83 & \multirow{2}{*}{187.774} \\*
 & H & 76 & 29.46 &  & 72 & 32 &  & 74 & 64 &  & \textbf{84} & 16 &  & 78 & 31.98 &  \\
\multirow{2}{*}{\textit{Stack Bowls Two}} & E & \textbf{96} & 36.64 & \multirow{2}{*}{129.979} & 90 & 32 & \multirow{2}{*}{115.667} & 92 & 64 & \multirow{2}{*}{107.338} & 94 & 16 & \multirow{2}{*}{139.929} & 90 & 30.03 & \multirow{2}{*}{119.606} \\*
 & H & 92 & 29.10 &  & 90 & 32 &  & 92 & 64 &  & \textbf{98} & 16 &  & 90 & 30.09 &  \\
\multirow{2}{*}{\textit{Stamp Seal}} & E & \textbf{100} & 29.97 & \multirow{2}{*}{59.814} & 78 & 32 & \multirow{2}{*}{54.773} & 76 & 64 & \multirow{2}{*}{49.932} & 98 & 16 & \multirow{2}{*}{64.284} & 90 & 27.15 & \multirow{2}{*}{63.290} \\*
 & H & 74 & 28.50 &  & 80 & 32 &  & \textbf{84} & 64 &  & \textbf{84} & 16 &  & 66 & 27.10 &  \\
\multirow{2}{*}{\textit{Turn Switch}} & E & 80 & 29.03 & \multirow{2}{*}{41.575} & \textbf{86} & 32 & \multirow{2}{*}{29.155} & 66 & 64 & \multirow{2}{*}{30.943} & 72 & 16 & \multirow{2}{*}{37.008} & 60 & 26.67 & \multirow{2}{*}{55.941} \\*
 & H & 52 & 28.27 &  & \textbf{66} & 32 &  & 64 & 64 &  & 56 & 16 &  & 60 & 26.40 &  \\
\midrule
\multirow{2}{*}{Average (50 tasks)} & E & 92.76 & 31.55 & \multirow{2}{*}{91.425} & 85.64 & 32.00 & \multirow{2}{*}{80.308} & 82.76 & 64.00 & \multirow{2}{*}{76.661} & \textbf{92.80} & 16.00 & \multirow{2}{*}{97.973} & 85.32 & 27.61 & \multirow{2}{*}{86.761} \\*
 & H & 89.20 & 28.41 &  & 83.68 & 32.00 &  & 83.20 & 64.00 &  & \textbf{90.68} & 16.00 &  & 82.08 & 27.52 &  \\
\bottomrule
\end{longtable}
\endgroup

% Required packages: booktabs, multirow, adjustbox
\clearpage
\begingroup
\scriptsize
\setlength{\tabcolsep}{1.45pt}
\setlength{\LTcapwidth}{\linewidth}
\setlength{\LTleft}{-0.45in plus 1fill}
\setlength{\LTright}{-0.45in plus 1fill}
\newcommand{\resulttablehead}{%
\toprule
\multirow{2}{*}{Task} & \multirow{2}{*}{Setting} & \multicolumn{3}{c}{Ours} & \multicolumn{3}{c}{RTC} & \multicolumn{3}{c}{Sync-12} & \multicolumn{3}{c}{AAC} \\
\cmidrule(lr){3-5}\cmidrule(lr){6-8}\cmidrule(lr){9-11}\cmidrule(lr){12-14}
 &  & SR $\uparrow$ & Chunk & Time (s) & SR $\uparrow$ & Chunk & Time (s) & SR $\uparrow$ & Chunk & Time (s) & SR $\uparrow$ & Chunk & Time (s) \\
\midrule
}
\begin{longtable}{@{}l c *{4}{r r r}@{}}
\caption{Per-task results of FastWAM on RoboTwin. E and H denote the clean (easy) and random (hard) settings, respectively. Time is the task-average successful-episode latency.}\label{tab:fastwam-robotwin-clean-random}\\
\resulttablehead
\endfirsthead
\multicolumn{1}{@{}l}{\tablename~\thetable\ (continued)} \\
\resulttablehead
\endhead
\multirow{2}{*}{\textit{Adjust Bottle}} & E & \textbf{100} & 19.29 & \multirow{2}{*}{27.010} & \textbf{100} & 18.77 & \multirow{2}{*}{24.380} & \textbf{100} & 12.00 & \multirow{2}{*}{39.660} & \textbf{100} & 17.23 & \multirow{2}{*}{28.930} \\*
 & H & \textbf{100} & 18.36 &  & \textbf{100} & 18.74 &  & \textbf{100} & 12.00 &  & \textbf{100} & 15.99 &  \\
\multirow{2}{*}{\textit{Beat Block Hammer}} & E & \textbf{100} & 17.94 & \multirow{2}{*}{28.250} & 98 & 17.83 & \multirow{2}{*}{22.290} & 98 & 12.00 & \multirow{2}{*}{38.840} & 98 & 16.85 & \multirow{2}{*}{28.000} \\*
 & H & \textbf{100} & 18.40 &  & 96 & 18.31 &  & 96 & 12.00 &  & \textbf{100} & 16.20 &  \\
\multirow{2}{*}{\textit{Blocks Ranking RGB}} & E & \textbf{100} & 19.74 & \multirow{2}{*}{69.840} & 92 & 19.77 & \multirow{2}{*}{62.430} & \textbf{100} & 12.00 & \multirow{2}{*}{75.840} & \textbf{100} & 16.34 & \multirow{2}{*}{84.240} \\*
 & H & \textbf{100} & 18.82 &  & \textbf{100} & 19.65 &  & \textbf{100} & 12.00 &  & 94 & 15.97 &  \\
\multirow{2}{*}{\textit{Blocks Ranking Size}} & E & 92 & 19.82 & \multirow{2}{*}{83.190} & 92 & 19.78 & \multirow{2}{*}{63.810} & \textbf{96} & 12.00 & \multirow{2}{*}{78.170} & 94 & 16.30 & \multirow{2}{*}{89.900} \\*
 & H & \textbf{96} & 18.78 &  & 90 & 19.77 &  & 94 & 12.00 &  & \textbf{96} & 16.06 &  \\
\multirow{2}{*}{\textit{Click Alarmclock}} & E & \textbf{100} & 16.06 & \multirow{2}{*}{17.580} & \textbf{100} & 18.60 & \multirow{2}{*}{14.080} & \textbf{100} & 12.00 & \multirow{2}{*}{31.430} & \textbf{100} & 16.48 & \multirow{2}{*}{17.300} \\*
 & H & \textbf{100} & 17.24 &  & \textbf{100} & 18.08 &  & \textbf{100} & 12.00 &  & \textbf{100} & 14.53 &  \\
\multirow{2}{*}{\textit{Click Bell}} & E & \textbf{100} & 17.66 & \multirow{2}{*}{16.570} & \textbf{100} & 18.22 & \multirow{2}{*}{13.570} & \textbf{100} & 12.00 & \multirow{2}{*}{31.250} & \textbf{100} & 17.58 & \multirow{2}{*}{16.820} \\*
 & H & \textbf{100} & 17.79 &  & 98 & 17.92 &  & \textbf{100} & 12.00 &  & \textbf{100} & 14.41 &  \\
\multirow{2}{*}{\textit{Dump Bin Bigbin}} & E & \textbf{96} & 18.84 & \multirow{2}{*}{36.450} & \textbf{96} & 19.08 & \multirow{2}{*}{33.630} & \textbf{96} & 12.00 & \multirow{2}{*}{50.580} & 92 & 16.21 & \multirow{2}{*}{45.200} \\*
 & H & 96 & 18.28 &  & 96 & 19.21 &  & \textbf{98} & 12.00 &  & \textbf{98} & 16.20 &  \\
\multirow{2}{*}{\textit{Grab Roller}} & E & \textbf{100} & 17.51 & \multirow{2}{*}{22.920} & \textbf{100} & 17.97 & \multirow{2}{*}{19.150} & \textbf{100} & 12.00 & \multirow{2}{*}{35.830} & \textbf{100} & 16.23 & \multirow{2}{*}{22.600} \\*
 & H & \textbf{100} & 16.79 &  & \textbf{100} & 17.17 &  & \textbf{100} & 12.00 &  & \textbf{100} & 15.54 &  \\
\multirow{2}{*}{\textit{Handover Block}} & E & 98 & 19.27 & \multirow{2}{*}{52.470} & \textbf{100} & 19.25 & \multirow{2}{*}{49.210} & 98 & 12.00 & \multirow{2}{*}{59.010} & 90 & 16.30 & \multirow{2}{*}{56.880} \\*
 & H & \textbf{88} & 18.70 &  & 78 & 19.65 &  & 82 & 12.00 &  & 86 & 16.14 &  \\
\multirow{2}{*}{\textit{Handover Mic}} & E & 98 & 18.88 & \multirow{2}{*}{48.700} & \textbf{100} & 19.13 & \multirow{2}{*}{38.800} & 98 & 12.00 & \multirow{2}{*}{55.720} & \textbf{100} & 16.30 & \multirow{2}{*}{49.250} \\*
 & H & \textbf{100} & 18.19 &  & 98 & 19.10 &  & \textbf{100} & 12.00 &  & \textbf{100} & 15.68 &  \\
\multirow{2}{*}{\textit{Hanging Mug}} & E & 58 & 19.52 & \multirow{2}{*}{63.670} & 54 & 19.62 & \multirow{2}{*}{61.630} & \textbf{66} & 12.00 & \multirow{2}{*}{68.860} & 56 & 16.18 & \multirow{2}{*}{68.130} \\*
 & H & 68 & 18.62 &  & 58 & 19.55 &  & 62 & 12.00 &  & \textbf{70} & 16.00 &  \\
\multirow{2}{*}{\textit{Lift Pot}} & E & \textbf{100} & 17.89 & \multirow{2}{*}{30.110} & \textbf{100} & 17.76 & \multirow{2}{*}{27.810} & \textbf{100} & 12.00 & \multirow{2}{*}{40.110} & \textbf{100} & 17.03 & \multirow{2}{*}{28.250} \\*
 & H & \textbf{100} & 17.32 &  & \textbf{100} & 17.55 &  & \textbf{100} & 12.00 &  & \textbf{100} & 15.14 &  \\
\multirow{2}{*}{\textit{Move Can Pot}} & E & 90 & 19.53 & \multirow{2}{*}{34.230} & 72 & 19.28 & \multirow{2}{*}{26.590} & \textbf{96} & 12.00 & \multirow{2}{*}{43.140} & 92 & 16.42 & \multirow{2}{*}{33.990} \\*
 & H & 94 & 18.26 &  & 82 & 19.02 &  & \textbf{98} & 12.00 &  & 92 & 15.78 &  \\
\multirow{2}{*}{\textit{Move Pillbottle Pad}} & E & \textbf{100} & 19.22 & \multirow{2}{*}{30.050} & 98 & 18.93 & \multirow{2}{*}{22.610} & 98 & 12.00 & \multirow{2}{*}{41.110} & \textbf{100} & 16.57 & \multirow{2}{*}{31.510} \\*
 & H & 98 & 18.35 &  & 98 & 19.01 &  & 96 & 12.00 &  & \textbf{100} & 15.96 &  \\
\multirow{2}{*}{\textit{Move Playingcard Away}} & E & \textbf{100} & 17.77 & \multirow{2}{*}{24.960} & 98 & 17.67 & \multirow{2}{*}{19.030} & \textbf{100} & 12.00 & \multirow{2}{*}{37.270} & \textbf{100} & 16.81 & \multirow{2}{*}{27.040} \\*
 & H & 98 & 18.07 &  & \textbf{100} & 17.52 &  & \textbf{100} & 12.00 &  & \textbf{100} & 16.46 &  \\
\multirow{2}{*}{\textit{Move Stapler Pad}} & E & \textbf{90} & 18.73 & \multirow{2}{*}{34.020} & 78 & 19.25 & \multirow{2}{*}{28.090} & 76 & 12.00 & \multirow{2}{*}{42.810} & 86 & 16.12 & \multirow{2}{*}{34.360} \\*
 & H & \textbf{78} & 18.40 &  & 76 & 19.24 &  & 74 & 12.00 &  & 68 & 15.94 &  \\
\multirow{2}{*}{\textit{Open Laptop}} & E & 98 & 18.03 & \multirow{2}{*}{33.260} & 98 & 17.95 & \multirow{2}{*}{26.690} & 98 & 12.00 & \multirow{2}{*}{43.330} & \textbf{100} & 16.10 & \multirow{2}{*}{31.360} \\*
 & H & \textbf{100} & 17.47 &  & \textbf{100} & 17.77 &  & \textbf{100} & 12.00 &  & \textbf{100} & 15.53 &  \\
\multirow{2}{*}{\textit{Open Microwave}} & E & 82 & 19.54 & \multirow{2}{*}{87.690} & \textbf{88} & 19.52 & \multirow{2}{*}{65.290} & 80 & 12.00 & \multirow{2}{*}{65.840} & 80 & 16.09 & \multirow{2}{*}{86.590} \\*
 & H & 58 & 18.89 &  & 48 & 19.86 &  & 58 & 12.00 &  & \textbf{62} & 15.93 &  \\
\multirow{2}{*}{\textit{Pick Diverse Bottles}} & E & 78 & 18.99 & \multirow{2}{*}{26.860} & \textbf{82} & 19.19 & \multirow{2}{*}{21.700} & 78 & 12.00 & \multirow{2}{*}{38.830} & 74 & 16.29 & \multirow{2}{*}{29.260} \\*
 & H & 84 & 18.11 &  & 88 & 18.89 &  & 88 & 12.00 &  & \textbf{90} & 16.05 &  \\
\multirow{2}{*}{\textit{Pick Dual Bottles}} & E & 98 & 18.28 & \multirow{2}{*}{24.610} & \textbf{100} & 18.50 & \multirow{2}{*}{20.730} & \textbf{100} & 12.00 & \multirow{2}{*}{38.930} & 98 & 16.46 & \multirow{2}{*}{27.200} \\*
 & H & \textbf{100} & 17.47 &  & 94 & 18.84 &  & 96 & 12.00 &  & 98 & 15.91 &  \\
\multirow{2}{*}{\textit{Place A2B Left}} & E & \textbf{96} & 19.41 & \multirow{2}{*}{31.200} & 94 & 19.20 & \multirow{2}{*}{23.810} & 94 & 12.00 & \multirow{2}{*}{41.610} & \textbf{96} & 16.90 & \multirow{2}{*}{31.510} \\*
 & H & 96 & 18.56 &  & 94 & 19.14 &  & 88 & 12.00 &  & \textbf{98} & 15.86 &  \\
\multirow{2}{*}{\textit{Place A2B Right}} & E & \textbf{98} & 19.33 & \multirow{2}{*}{30.900} & 94 & 19.33 & \multirow{2}{*}{25.030} & 96 & 12.00 & \multirow{2}{*}{41.900} & \textbf{98} & 16.61 & \multirow{2}{*}{32.450} \\*
 & H & 92 & 18.57 &  & 94 & 19.16 &  & \textbf{98} & 12.00 &  & 96 & 15.85 &  \\
\multirow{2}{*}{\textit{Place Bread Basket}} & E & \textbf{94} & 19.49 & \multirow{2}{*}{46.680} & 92 & 19.44 & \multirow{2}{*}{38.500} & 90 & 12.00 & \multirow{2}{*}{52.440} & 90 & 16.08 & \multirow{2}{*}{46.670} \\*
 & H & 96 & 18.25 &  & 92 & 19.39 &  & \textbf{98} & 12.00 &  & 92 & 15.92 &  \\
\multirow{2}{*}{\textit{Place Bread Skillet}} & E & 92 & 18.96 & \multirow{2}{*}{34.920} & 92 & 18.80 & \multirow{2}{*}{27.800} & \textbf{94} & 12.00 & \multirow{2}{*}{44.540} & 90 & 16.51 & \multirow{2}{*}{35.210} \\*
 & H & \textbf{96} & 18.04 &  & \textbf{96} & 18.75 &  & 86 & 12.00 &  & \textbf{96} & 16.08 &  \\
\multirow{2}{*}{\textit{Place Burger Fries}} & E & 96 & 19.08 & \multirow{2}{*}{50.860} & 94 & 19.15 & \multirow{2}{*}{41.630} & \textbf{98} & 12.00 & \multirow{2}{*}{56.620} & \textbf{98} & 16.40 & \multirow{2}{*}{51.020} \\*
 & H & \textbf{100} & 18.29 &  & 96 & 19.16 &  & \textbf{100} & 12.00 &  & \textbf{100} & 15.86 &  \\
\multirow{2}{*}{\textit{Place Can Basket}} & E & 68 & 19.41 & \multirow{2}{*}{63.970} & 68 & 19.22 & \multirow{2}{*}{57.410} & \textbf{70} & 12.00 & \multirow{2}{*}{65.360} & 60 & 16.14 & \multirow{2}{*}{63.410} \\*
 & H & 74 & 18.57 &  & \textbf{80} & 19.25 &  & 64 & 12.00 &  & 56 & 15.89 &  \\
\multirow{2}{*}{\textit{Place Cans Plasticbox}} & E & \textbf{100} & 18.79 & \multirow{2}{*}{54.670} & 98 & 18.49 & \multirow{2}{*}{45.110} & 96 & 12.00 & \multirow{2}{*}{58.530} & 94 & 16.17 & \multirow{2}{*}{53.250} \\*
 & H & \textbf{100} & 18.31 &  & 98 & 18.71 &  & \textbf{100} & 12.00 &  & 98 & 15.94 &  \\
\multirow{2}{*}{\textit{Place Container Plate}} & E & 96 & 19.70 & \multirow{2}{*}{23.010} & \textbf{100} & 19.16 & \multirow{2}{*}{24.090} & \textbf{100} & 12.00 & \multirow{2}{*}{41.370} & 98 & 17.35 & \multirow{2}{*}{31.020} \\*
 & H & 98 & 18.69 &  & 98 & 18.99 &  & \textbf{100} & 12.00 &  & \textbf{100} & 15.70 &  \\
\multirow{2}{*}{\textit{Place Dual Shoes}} & E & 92 & 19.50 & \multirow{2}{*}{55.080} & \textbf{94} & 19.32 & \multirow{2}{*}{44.750} & 78 & 12.00 & \multirow{2}{*}{59.020} & 90 & 16.31 & \multirow{2}{*}{55.910} \\*
 & H & \textbf{86} & 18.55 &  & \textbf{86} & 19.37 &  & 84 & 12.00 &  & 82 & 15.95 &  \\
\multirow{2}{*}{\textit{Place Empty Cup}} & E & \textbf{100} & 18.87 & \multirow{2}{*}{32.560} & \textbf{100} & 18.49 & \multirow{2}{*}{26.040} & \textbf{100} & 12.00 & \multirow{2}{*}{43.280} & \textbf{100} & 16.84 & \multirow{2}{*}{32.990} \\*
 & H & \textbf{100} & 18.51 &  & 98 & 18.60 &  & \textbf{100} & 12.00 &  & \textbf{100} & 16.31 &  \\
\multirow{2}{*}{\textit{Place Fan}} & E & \textbf{98} & 18.95 & \multirow{2}{*}{25.510} & 92 & 19.13 & \multirow{2}{*}{27.780} & 92 & 12.00 & \multirow{2}{*}{44.680} & 96 & 16.63 & \multirow{2}{*}{35.390} \\*
 & H & 96 & 18.39 &  & \textbf{98} & 18.89 &  & 94 & 12.00 &  & \textbf{98} & 16.18 &  \\
\multirow{2}{*}{\textit{Place Mouse Pad}} & E & 86 & 19.24 & \multirow{2}{*}{31.730} & 86 & 19.16 & \multirow{2}{*}{26.200} & 84 & 12.00 & \multirow{2}{*}{41.850} & \textbf{88} & 16.60 & \multirow{2}{*}{33.430} \\*
 & H & \textbf{92} & 18.54 &  & 82 & 19.22 &  & 84 & 12.00 &  & 88 & 15.83 &  \\
\multirow{2}{*}{\textit{Place Object Basket}} & E & \textbf{90} & 19.33 & \multirow{2}{*}{63.690} & 76 & 19.56 & \multirow{2}{*}{53.620} & 88 & 12.00 & \multirow{2}{*}{68.640} & 82 & 16.25 & \multirow{2}{*}{65.740} \\*
 & H & 80 & 18.67 &  & 82 & 19.58 &  & 78 & 12.00 &  & \textbf{86} & 15.91 &  \\
\multirow{2}{*}{\textit{Place Object Scale}} & E & 94 & 19.44 & \multirow{2}{*}{30.720} & \textbf{96} & 19.13 & \multirow{2}{*}{24.920} & \textbf{96} & 12.00 & \multirow{2}{*}{41.090} & 90 & 16.80 & \multirow{2}{*}{30.020} \\*
 & H & \textbf{98} & 18.29 &  & 90 & 19.22 &  & 92 & 12.00 &  & 96 & 15.95 &  \\
\multirow{2}{*}{\textit{Place Object Stand}} & E & 92 & 18.90 & \multirow{2}{*}{28.300} & 88 & 18.76 & \multirow{2}{*}{23.310} & \textbf{96} & 12.00 & \multirow{2}{*}{40.500} & \textbf{96} & 17.14 & \multirow{2}{*}{29.570} \\*
 & H & 90 & 18.73 &  & \textbf{96} & 18.82 &  & 94 & 12.00 &  & 90 & 15.75 &  \\
\multirow{2}{*}{\textit{Place Phone Stand}} & E & \textbf{98} & 19.11 & \multirow{2}{*}{27.240} & 96 & 19.43 & \multirow{2}{*}{22.000} & 94 & 12.00 & \multirow{2}{*}{38.400} & 96 & 16.16 & \multirow{2}{*}{27.270} \\*
 & H & 96 & 17.75 &  & \textbf{100} & 19.31 &  & 94 & 12.00 &  & 98 & 15.82 &  \\
\multirow{2}{*}{\textit{Place Shoe}} & E & 94 & 18.79 & \multirow{2}{*}{34.580} & 94 & 19.36 & \multirow{2}{*}{31.090} & \textbf{96} & 12.00 & \multirow{2}{*}{44.210} & \textbf{96} & 16.18 & \multirow{2}{*}{34.480} \\*
 & H & \textbf{98} & 18.08 &  & 96 & 19.40 &  & \textbf{98} & 12.00 &  & \textbf{98} & 15.98 &  \\
\multirow{2}{*}{\textit{Press Stapler}} & E & 96 & 17.35 & \multirow{2}{*}{22.320} & 96 & 17.37 & \multirow{2}{*}{20.180} & \textbf{98} & 12.00 & \multirow{2}{*}{35.820} & \textbf{98} & 16.23 & \multirow{2}{*}{22.200} \\*
 & H & \textbf{100} & 16.87 &  & \textbf{100} & 16.80 &  & \textbf{100} & 12.00 &  & \textbf{100} & 15.23 &  \\
\multirow{2}{*}{\textit{Put Bottles Dustbin}} & E & 86 & 19.88 & \multirow{2}{*}{83.050} & 84 & 19.78 & \multirow{2}{*}{75.650} & \textbf{90} & 12.00 & \multirow{2}{*}{92.190} & 84 & 16.11 & \multirow{2}{*}{106.360} \\*
 & H & \textbf{88} & 18.78 &  & 78 & 19.84 &  & 86 & 12.00 &  & 86 & 16.09 &  \\
\multirow{2}{*}{\textit{Put Object Cabinet}} & E & \textbf{94} & 19.48 & \multirow{2}{*}{56.200} & 90 & 19.50 & \multirow{2}{*}{52.270} & 88 & 12.00 & \multirow{2}{*}{63.850} & 90 & 16.55 & \multirow{2}{*}{60.860} \\*
 & H & \textbf{92} & 18.61 &  & 80 & 19.64 &  & 86 & 12.00 &  & 90 & 15.95 &  \\
\multirow{2}{*}{\textit{Rotate QRcode}} & E & 90 & 19.48 & \multirow{2}{*}{25.200} & \textbf{92} & 19.09 & \multirow{2}{*}{27.340} & 90 & 12.00 & \multirow{2}{*}{43.380} & 90 & 16.55 & \multirow{2}{*}{35.670} \\*
 & H & 90 & 18.46 &  & \textbf{94} & 19.18 &  & 92 & 12.00 &  & 84 & 15.90 &  \\
\multirow{2}{*}{\textit{Scan Object}} & E & \textbf{98} & 19.23 & \multirow{2}{*}{38.380} & 92 & 19.27 & \multirow{2}{*}{35.870} & 92 & 12.00 & \multirow{2}{*}{50.620} & 94 & 16.53 & \multirow{2}{*}{40.880} \\*
 & H & 90 & 18.47 &  & 92 & 19.23 &  & \textbf{98} & 12.00 &  & 92 & 16.15 &  \\
\multirow{2}{*}{\textit{Shake Bottle}} & E & \textbf{100} & 18.40 & \multirow{2}{*}{22.870} & \textbf{100} & 18.30 & \multirow{2}{*}{17.390} & \textbf{100} & 12.00 & \multirow{2}{*}{36.170} & \textbf{100} & 16.85 & \multirow{2}{*}{24.300} \\*
 & H & \textbf{100} & 17.76 &  & \textbf{100} & 18.14 &  & \textbf{100} & 12.00 &  & \textbf{100} & 16.70 &  \\
\multirow{2}{*}{\textit{Shake Bottle Horizontally}} & E & \textbf{100} & 18.35 & \multirow{2}{*}{22.900} & \textbf{100} & 18.21 & \multirow{2}{*}{19.200} & \textbf{100} & 12.00 & \multirow{2}{*}{36.100} & \textbf{100} & 16.87 & \multirow{2}{*}{22.360} \\*
 & H & \textbf{100} & 17.79 &  & \textbf{100} & 18.10 &  & \textbf{100} & 12.00 &  & \textbf{100} & 16.63 &  \\
\multirow{2}{*}{\textit{Stack Blocks Three}} & E & 96 & 19.68 & \multirow{2}{*}{81.210} & \textbf{100} & 19.63 & \multirow{2}{*}{71.350} & \textbf{100} & 12.00 & \multirow{2}{*}{76.600} & \textbf{100} & 16.33 & \multirow{2}{*}{86.220} \\*
 & H & 94 & 18.81 &  & 98 & 19.62 &  & \textbf{100} & 12.00 &  & 98 & 16.00 &  \\
\multirow{2}{*}{\textit{Stack Blocks Two}} & E & \textbf{100} & 19.47 & \multirow{2}{*}{55.460} & \textbf{100} & 19.31 & \multirow{2}{*}{43.390} & \textbf{100} & 12.00 & \multirow{2}{*}{58.820} & \textbf{100} & 16.33 & \multirow{2}{*}{57.960} \\*
 & H & \textbf{100} & 18.68 &  & 98 & 19.40 &  & \textbf{100} & 12.00 &  & \textbf{100} & 16.05 &  \\
\multirow{2}{*}{\textit{Stack Bowls Three}} & E & 76 & 19.88 & \multirow{2}{*}{80.870} & 66 & 19.79 & \multirow{2}{*}{78.730} & 76 & 12.00 & \multirow{2}{*}{83.230} & \textbf{78} & 16.28 & \multirow{2}{*}{93.050} \\*
 & H & \textbf{84} & 18.93 &  & \textbf{84} & 19.75 &  & 80 & 12.00 &  & 62 & 16.07 &  \\
\multirow{2}{*}{\textit{Stack Bowls Two}} & E & 86 & 19.82 & \multirow{2}{*}{56.840} & 84 & 19.51 & \multirow{2}{*}{45.320} & 92 & 12.00 & \multirow{2}{*}{60.490} & \textbf{94} & 16.49 & \multirow{2}{*}{62.870} \\*
 & H & \textbf{98} & 18.79 &  & \textbf{98} & 19.39 &  & \textbf{98} & 12.00 &  & 92 & 15.95 &  \\
\multirow{2}{*}{\textit{Stamp Seal}} & E & 82 & 19.18 & \multirow{2}{*}{28.420} & \textbf{88} & 18.99 & \multirow{2}{*}{22.080} & \textbf{88} & 12.00 & \multirow{2}{*}{39.670} & \textbf{88} & 16.76 & \multirow{2}{*}{30.640} \\*
 & H & 92 & 18.49 &  & 88 & 18.87 &  & \textbf{94} & 12.00 &  & \textbf{94} & 15.80 &  \\
\multirow{2}{*}{\textit{Turn Switch}} & E & \textbf{60} & 19.26 & \multirow{2}{*}{19.430} & 54 & 19.14 & \multirow{2}{*}{15.290} & 58 & 12.00 & \multirow{2}{*}{31.730} & 52 & 16.30 & \multirow{2}{*}{16.400} \\*
 & H & 62 & 18.64 &  & 62 & 19.37 &  & \textbf{66} & 12.00 &  & 64 & 15.84 &  \\
\midrule
\multirow{2}{*}{Average (50 tasks)} & E & \textbf{92.52} & 18.97 & \multirow{2}{*}{41.024} & 90.92 & 18.98 & \multirow{2}{*}{34.973} & 92.24 & 12.00 & \multirow{2}{*}{49.785} & 91.72 & 16.50 & \multirow{2}{*}{43.718} \\*
 & H & \textbf{92.72} & 18.28 &  & 90.96 & 18.94 &  & 91.88 & 12.00 &  & 91.76 & 15.89 &  \\
\bottomrule
\end{longtable}
\endgroup

\FloatBarrier

\end{document}